\documentclass[lettersize,journal]{IEEEtran}
\usepackage{amsmath,amsfonts}
\usepackage{algorithmic}
\usepackage{algorithm}
\usepackage{array}
\usepackage[caption=false,font=normalsize,labelfont=sf,textfont=sf]{subfig}
\usepackage{textcomp}
\usepackage{stfloats}
\usepackage{url}
\usepackage{verbatim}
\usepackage{graphicx}
\usepackage{cite}
\usepackage{enumitem}
\usepackage{hyperref}
\usepackage{xcolor}
\begin{document}

\title{Aerial Vision-Language Navigation with a Unified Framework \\ for Spatial, Temporal and Embodied Reasoning}

\author{
Huilin Xu,~\IEEEmembership{Graduate Student Member,~IEEE},
Zhuoyang Liu,~\IEEEmembership{Graduate Student Member,~IEEE}, \\
Yixiang Luomei,~\IEEEmembership{Member,~IEEE}, and 
Feng Xu,~\IEEEmembership{Senior Member,~IEEE}
\thanks{The authors are with the Key Laboratory for Information Science of Electromagnetic Waves (Ministry of Education), School of Information Science and Technology, Fudan University, Shanghai 200433, China (e-mail:
fengxu@fudan.edu.cn).}
}


\markboth{Journal of \LaTeX\ Class Files,~Vol.~14, No.~8, August~2021}%
{Shell \MakeLowercase{\textit{et al.}}: A Sample Article Using IEEEtran.cls for IEEE Journals}


\maketitle

\begin{abstract}
Aerial Vision-and-Language Navigation (VLN) aims to enable unmanned aerial vehicles (UAVs) to interpret natural language instructions and navigate complex urban environments using onboard visual observations. This task holds promise for real-world applications such as low-altitude inspection, search-and-rescue, and autonomous aerial delivery. Existing methods often rely on panoramic images, depth information, or odometry to support spatial reasoning and action planning. Such reliance on additional sensing modalities increases system cost and integration complexity. In this work, we present a unified aerial VLN framework that operates solely on egocentric monocular RGB observations and natural language instructions. The model formulates navigation as a next-token prediction problem, jointly optimizing spatial perception, trajectory reasoning, and action prediction through prompt-guided multi-task learning.  Moreover, we propose a keyframe selection strategy to reduce visual redundancy by retaining semantically informative frames, along with an action merging and label reweighting mechanism that mitigates long-tailed supervision imbalance and facilitates stable multi-task co-training. Extensive experiments on the AerialVLN and OpenFly benchmarks validate the effectiveness of our method. Under the challenging monocular RGB-only setting, our model achieves strong results across both seen and unseen environments. It significantly outperforms existing monocular baselines and narrows the performance gap with state-of-the-art panoramic RGB-D counterparts. Comprehensive ablation studies further validate the contribution of our architectural design and choices. Our code is publicly available at \url{https://github.com/return-sleep/AeroAct}.

\end{abstract}


\section{Introduction}
\IEEEPARstart{U}{nmanned} Aerial Vehicle (UAV) has become an indispensable tool in modern remote sensing applications, playing a central role in infrastructure inspection, environmental monitoring,  and emergency response \cite{khan2022uavsurvey}. Previous research has largely focused on passive perception tasks, including object detection \cite{leng2022pareto, chen2023high} and tracking \cite{chen2023cross, wu2024temporal} from aerial images or videos, without interaction with the world. In contrast, aerial navigation tasks require the drone to perceive, reason, and act in dynamic environments.  Recently, aerial Vision-and-Language Navigation (VLN) \cite{liu2023aerialvln} has emerged as a new paradigm, where drones follow high-level language instructions to navigate to the destination through 3D outdoor environments.  By leveraging natural language as a human-centric interface, aerial VLN significantly reduces the reliance on expert pilots, lowers the barrier of human-UAV interaction, and enables intuitive task specification in high-stakes scenarios.

\begin{figure}[t!]
    \centering
    \includegraphics[width=\linewidth]{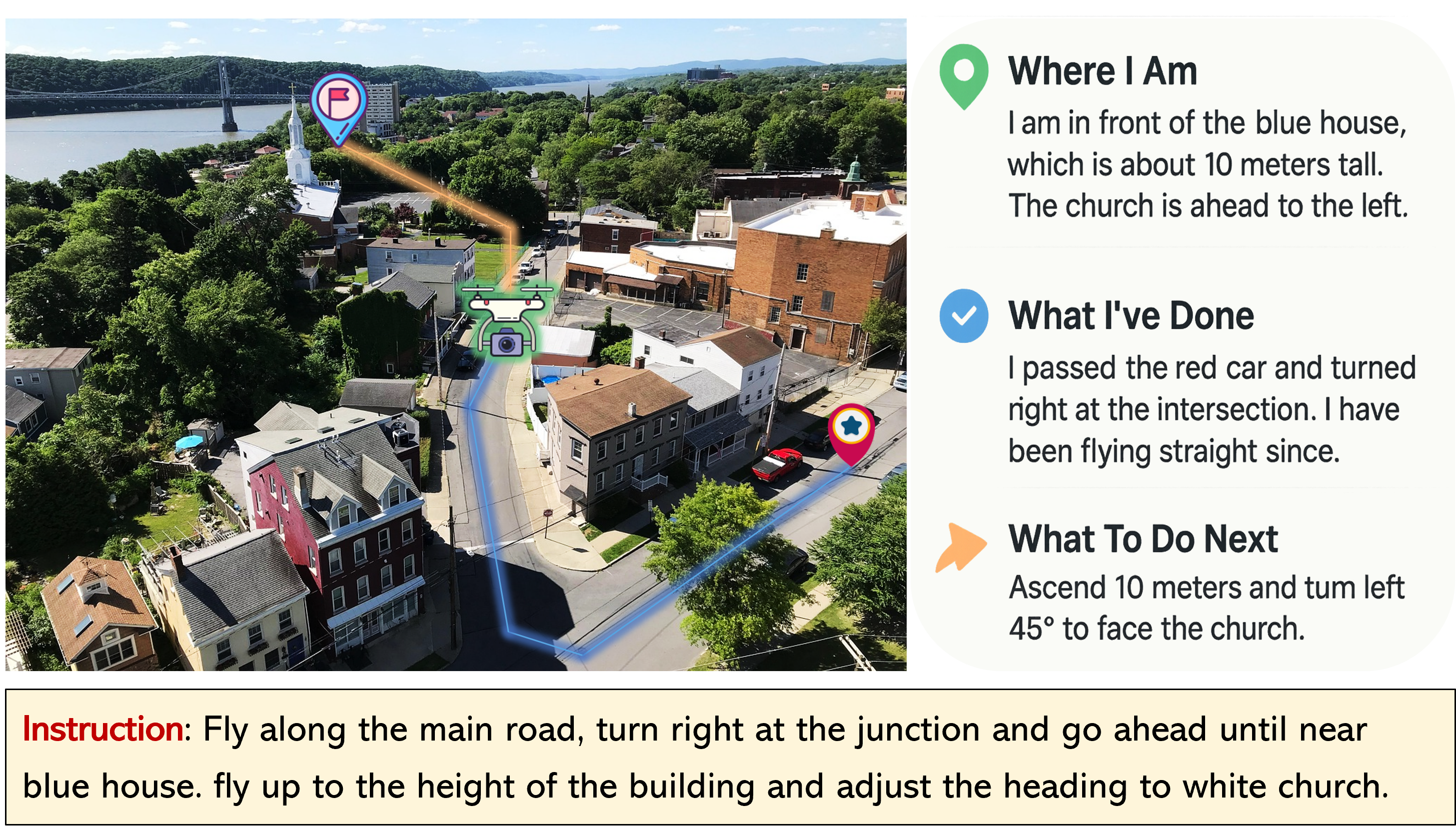}
    \caption{
    Aerial vision-language navigation. Left: A drone receives a natural-language instruction along with egocentric visual observations and is required to navigate to the destination in a complex outdoor environment. Right: This task relies on the agent’s ability to maintain an accurate understanding of its navigational situation, including estimating its current position, interpreting its progress within the instruction, and determining the next movement consistent with the described route. The example highlights these dimensions of temporal and spatial reasoning, which are central to reliable long-horizon aerial navigation.
    }
    \label{fig:teaser}
\end{figure}

Recent works have explored the design of benchmarks and datasets to facilitate research in aerial vision-and-language navigation. AVDN \cite{fan2023AVDN} first proposed a dialogue-based setting involving asynchronous interactions between a human commander and a UAV agent. AerialVLN \cite{liu2023aerialvln} introduced high-fidelity city-scale simulations with diverse human-annotated trajectories. CityNav\cite{lee2025citynavdataset} extended this line by leveraging real-world urban reconstructions. OpenUAV \cite{wangopenuav} formulated aerial VLN as full-trajectory prediction with human-in-the-loop evaluation, and OpenFly \cite{gao2025openfly} scaled up scene and instruction diversity via automatic generation. AeroDuo \cite{wu2025aeroduo} explores collaborative multi-agent instruction following with altitude-aware role assignment. Moreover, CityEQA \cite{zhao2025cityeqa}  and 3D Open-EQA \cite{zhang2025open3dvqa} have proposed embodied question answering benchmarks to assess models’ perception and reasoning capabilities from an aerial perspective. However, as illustrated in Fig. \ref{fig:teaser}, aerial VLN requires an agent to continuously integrate current observation, trajectory history, and high-level instruction. Concretely, the agent must recognize its location from egocentric views, assess task progress relative to instruction, and infer the next action consistent with the intended route.  Aerial VLN presents unique challenges across three core aspects.

\begin{enumerate}
    \item \textit{Unified Vision-Language-Action Alignment:}
    UAVs struggle for high-dimensional action spaces involving both horizontal and vertical movements, where altitude changes affect visibility, orientation, and control feasibility. Mapping free-form natural language instructions to executable flight commands requires reliable alignment between egocentric semantics and spatial geometry.
    
    \item \textit{Large-scale Outdoor Aerial Scenes:}
    UAVs operate in large-scale urban environments with dense infrastructures, complex spatial layouts, and highly dynamic semantic distributions. This complexity demands precise spatial perception (\textit{"ascend to the height of the street lamp"}) and the ability to robustly ground language-referred landmarks (\textit{"the gray house with a slope"}) using egocentric observations.

    \item \textit{Long-Horizon Temporal Reasoning:}
    Aerial instructions often describe multi-stage flight plans that span long distances and evolving visual contexts. Agents must not only interpret sequential goals but also maintain trajectory-level awareness, track historical behavior, and align current decisions with global navigation intent.
   
\end{enumerate}

Massive endeavors have attempted to alleviate these challenges. One line of research leverages pre-trained vision-language models (VLMs) to enable zero-shot instruction following \cite{gao2024aerialstmr,zhang2025citynavagent,uav-on}. These approaches introduce explicit intermediate representations and memory mechanisms to mitigate large-scale spatial reasoning difficulties and preserve long-term spatial context. Large language models are subsequently employed for high-level planning. However, such modular designs decouple perception, representation, and decision-making into separate stages, which may introduce error accumulation and restrict joint optimization across complementary navigation capabilities. Another line of research adopts learning-based end-to-end modeling to directly map visual observations and language instructions to embodied action sequences \cite{liu2023aerialvln,zhao2025aerialviewselection,lee2025citynavdataset}. These methods enhance cross-modal alignment by incorporating geographic spatial representations or history-aware fusion modules to better utilize spatial cues and historical observations. More recent efforts further leverage large vision-language models with strong multimodal capabilities to improve reasoning over aerial trajectories \cite{gao2025openfly, cai2025flightgpt,wu2025vla-an}. While these approaches have achieved promising performance, many rely on auxiliary inputs such as panoramic cameras, depth sensors, or odometry, which increase hardware complexity and system integration overhead. Moreover, the limited memory context of current models remains a bottleneck, making long-horizon aerial navigation particularly challenging.

To address the aforementioned limitations without introducing additional sensing modalities, while leveraging the strong multimodal capability of vision-language models, we develop a unified framework for aerial VLN that relies solely on onboard egocentric RGB observations and natural language instructions. The proposed framework directly predicts the next navigation action, bridging language semantics with visual context. We formulate navigation under a Next-Token Prediction (NTP) paradigm, jointly modeling spatial perception, trajectory reasoning, and action generation within a single vision-language backbone. This unified formulation enables cross-modal alignment and trajectory-level reasoning to be optimized under a shared objective without architectural modification. Within this framework, task-specific prompts are introduced to steer learning towards distinct reasoning objectives while preserving architectural simplicity. Furthermore, considering the structural characteristics of aerial navigation, including substantial visual redundancy during continuous flight and imbalance in action distributions, we incorporate an aerial-specific design consisting of keyframe selection, action merging, and label reweighting. These components are integrated into the overall learning framework to improve supervision balance and sequence modeling stability. Experimental results across multiple benchmarks demonstrate that our method consistently outperforms baseline models.

In summary, our paper makes the following contributions:
\begin{enumerate}
    \item \textit{Unified Next-Token Formulation:}
    We formulate aerial VLN as a next-token prediction problem, jointly modeling spatial perception, trajectory reasoning, and action generation within a single autoregressive backbone, thereby facilitating tighter cross-modal alignment.
    
    \item \textit{Prompt-Driven Multi-Task Supervision:}
    We introduce task-specific prompts to incorporate auxiliary objectives for spatial grounding and trajectory reasoning, strengthening the model’s reasoning ability over spatial structure and temporal evolution. 

    \item \textit{Aerial-Specific Design:}
    We propose an aerial-specific design and training strategy, including keyframe selection, action merging, and label reweighting, to address visual redundancy and action imbalance inherent in aerial navigation for stable training.
   
\end{enumerate}

\section{Related Works}

\subsection{Vision-and-Language Navigation}
Vision-and-language navigation (VLN) tasks require an agent to follow natural language instructions using visual observations to achieve a target location \cite{wu2024vln_survey, NavComposer}. Early VLN benchmarks formulated navigation on a predefined graph of panoramic viewpoints, where the agent selects discrete actions to move between connected nodes \cite{anderson2018r2r, ku2020rxr,2024vln}. To move beyond the discrete setting, VLN-CE \cite{krantz2020cma} replaced the navigation graph with free movement in simulation, where the agent executes mid-level actions, such as moving forward and turning. A variety of learning-based methods have been proposed under this setting, including cross-modal attention \cite{hong2021vln_bert}, memory-augmented agents \cite{wang2023gridmm}, and planning-based approaches \cite{an2024etpnav, wang2023dreamwalker, 2024planning}.
A growing line of work has focused on training vision-language models that directly map RGB observations and instructions to navigation actions. NaVid \cite{zhang2024navid} demonstrates that a vision-language model trained on large-scale continuous trajectories can effectively use monocular RGB video to model temporal context and predict actions, achieving strong performance. NaVILA \cite{cheng2025navila} introduces a two-stage framework that maps visual observations and instructions into mid-level language actions, which are then executed by a learned locomotion policy, enabling instruction-conditioned control in low-level continuous environments. MonoDream \cite{wang2025monodream} introduces auxiliary representation learning tasks that train a monocular agent to align visual semantics and action intent by predicting latent panoramic features. Uni-NaVid \cite{zhang2024uninavid} unifies diverse navigation tasks within a single end-to-end framework by learning from multi-task RGB video data. VLN-R1 \cite{qi2025vln} combines supervised and reinforcement fine-tuning to enhance temporal reasoning and multi-step action planning.

While these approaches have made substantial progress, they are primarily developed for ground-based agents in indoor environments and typically operate within limited spatial scales. Our work instead focuses on aerial VLN in large-scale urban environments, with inherent challenges, such as viewpoint variability, long-range flight paths, and complex 3D outdoor scenes. We adopt a unified framework and enhance it with prompt-guided auxiliary supervision to strengthen spatial perception and trajectory reasoning over long horizons.

\subsection{Aerial Vision-and-Language Navigation}

Aerial navigation has attracted growing attention, with early efforts focusing on conventional tasks such as obstacle avoidance \cite{lindqvist2021obstacleavoidance, zhang2025obstacleavoidancenature} and object tracking \cite{wang2019targettracking,savkin2023effective} based on onboard cameras or GPS signals. Recent advances explore language-guided aerial navigation, aiming to empower UAVs to interpret and execute natural language instructions in large-scale, open-world scenarios \cite{liu2023aerialvln}.  This task introduces additional challenges beyond visual navigation, including grounding language instructions in visual observations, reasoning over temporal context, and aligning multimodal information for robust decision-making. 

Research on aerial VLN has gradually evolved with the introduction of dedicated benchmarks and diverse solution paradigms. AVDN \cite{fan2023AVDN} introduces a dialogue-based UAV-VLN setting, while AerialVLN \cite{liu2023aerialvln} and CityNav \cite{lee2025citynavdataset} construct large-scale outdoor datasets using synthetic cities and real-world 3D reconstruction, respectively. OpenUAV \cite{wangopenuav} formulates UAV-VLN as full-trajectory prediction and incorporates human-in-the-loop evaluation. OpenFly \cite{gao2025openfly} scales up dataset diversity through automatic generation over varied high-fidelity virtual environments. Early approaches to aerial VLN adapted models from ground-based navigation, but often struggled with dynamic viewpoints and large-scale outdoor complexity. More recently, with the rise of large language models (LLMs) and vision-language models (VLMs), a line of work explores training-free paradigms for zero-shot aerial VLN. These methods convert visual observations into text prompts in the form of topological graphs, symbolic maps, or structured semantic matrices and rely on LLMs to infer the next navigation action or waypoint. Among them, STMR \cite{gao2024aerialstmr} introduces a Semantic-Topo-Metric Representation that encodes current view, pose, and subgoal cues into matrix-form prompts for spatial reasoning over landmarks and directions. CityNavAgent \cite{zhang2025citynavagent} further enhances semantic grounding via symbolic 2D landmark maps derived from real aerial imagery and spatial detection. In contrast, learning-based approaches directly align visual observations and language instructions through end-to-end optimization. LAG \cite{liu2023aerialvln} introduces a look-ahead guidance mechanism to Cross-Modal Attention method \cite{krantz2020cma}. Grid-based View Selection \cite{zhao2025aerialviewselection} reformulates aerial VLN as a grid-based view selection problem with coupled vertical-horizontal action modeling, bird’s-eye history fusion, and cross-modal alignment. More closely related to our work, OpenFly \cite{gao2025openfly} proposes keyframe-aware modeling and adaptive token merging to improve efficiency and stability in long-horizon navigation. 

However, many existing aerial VLN approaches rely on additional sensory inputs or explicit structural representations to facilitate navigation. Differently, we focus on an egocentric monocular RGB setting and formulate aerial VLN under a unified next-token prediction paradigm. Within this formulation, we incorporate two auxiliary objectives to enhance the scene understanding and temporal reasoning. We further introduce aerial-specific training strategies, including keyframe selection, action merging, and label reweighting, to reduce visual redundancy and mitigate label imbalance.

\begin{figure}[t!]
\color{black}
    \centering
    \includegraphics[width=\linewidth]{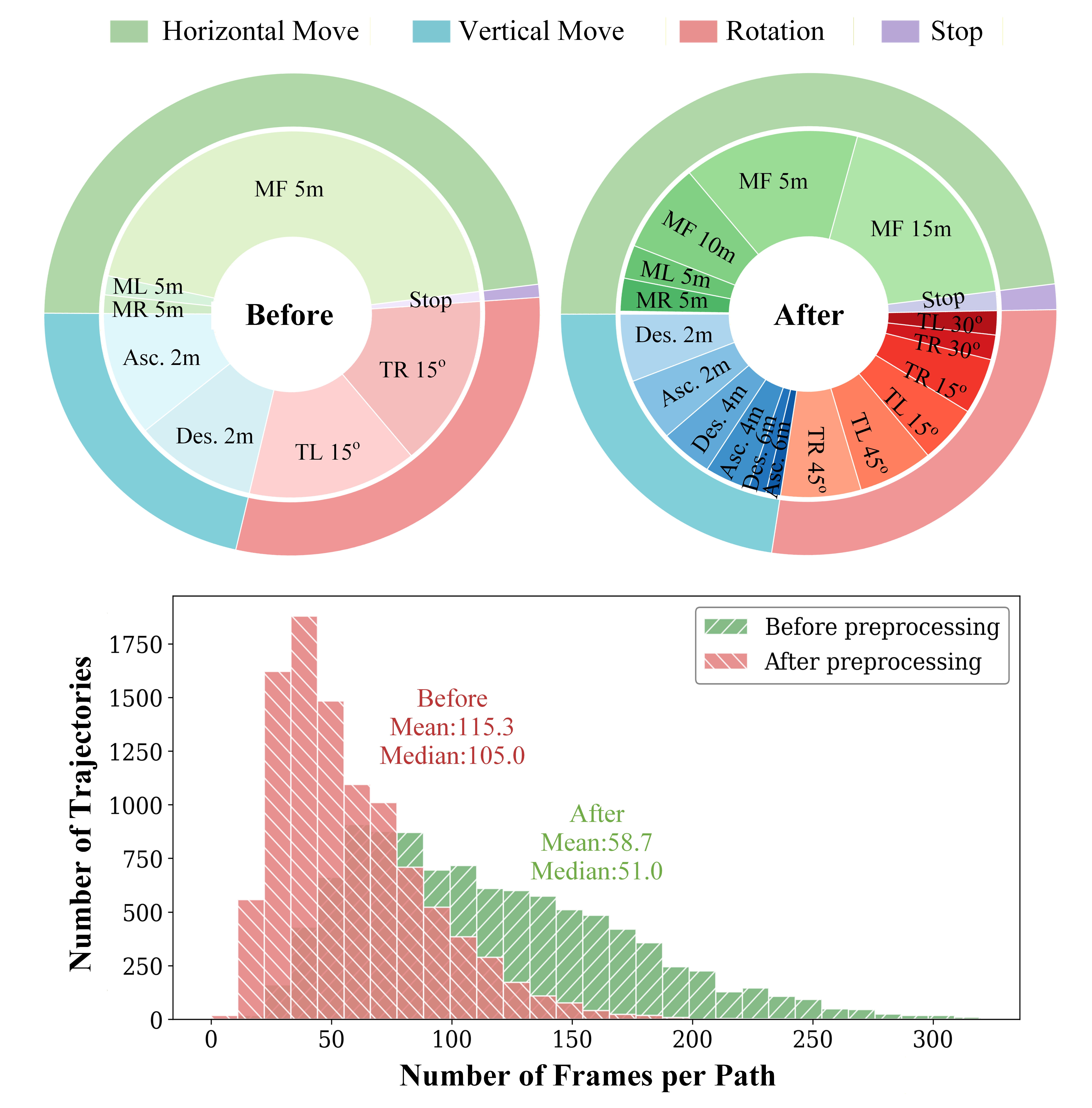}
    \caption{
    Trajectory statistics before and after data preprocessing, showing that action merging and keyframe selection yield a richer action space and more compact navigation sequences. For brevity, we use MF for \textit{move forward}, etc.
    }
    \label{fig:preprocess_vis}
\end{figure}

\section{Method}

\subsection{Task Definition}
\noindent \textbf{Aerial Vision-Language Navigation.}
The goal of aerial VLN is to learn a policy that enables drones to follow high-level language instructions to navigate through 3D outdoor environments and reach the target location. Specifically, an aerial agent is tasked with navigating a continuous 3D space according to a natural language instruction $I=\{w_1, w_2, \ldots, w_{N_I}\}$, which comprises $N_I$ words describing the desired flight route. At each time step $t$, the agent receives a visual trajectory $\tau_t = \{x_1, \ldots, x_t\}$, where $x_t \in \mathbb{R}^{H \times W \times 3}$ is the first-person RGB image captured by its onboard camera.
Conditioned on visual trajectory $\tau_t$ and the instruction $I$, the agent selects an action $a_t$ from a discrete action set 
$\mathcal{A}$. Executing $a_t$ moves the agent to the next state in the environment, yielding the subsequent image observation $x_{t+1}$. An episode terminates when the agent chooses the \textit{stop} action or reaches the maximum step limit. A navigation episode is considered successful if the agent's final position lies within $20$ meters of the destination. The action space and corresponding motion magnitudes are defined according to the specific environment settings.
In the AerialVLN environment, $\mathcal{A} = \{\textit{move forward},\allowbreak \textit{turn left},\allowbreak 
\textit{turn right},\allowbreak \textit{ascend},\allowbreak 
\textit{descend},\allowbreak \textit{move left},\allowbreak 
\textit{move right}, \allowbreak \textit{stop}\}$. Horizontal motions displace the agent by $5$ units, vertical motions by $2$ units, and turning applies a fixed $15^\circ$ change in orientation. In the OpenFly environment, $\mathcal{A} = \{\textit{move forward},\allowbreak \textit{turn left},\allowbreak 
\textit{turn right},\allowbreak \textit{ascend},\allowbreak 
\textit{descend},\allowbreak \textit{ stop}\}$. Horizontal motions displace the agent by $3$ units, vertical motions by $3$ units, and turning applies  $30^\circ$.

\begin{figure*}[t!]
    \centering
    \includegraphics[width= \textwidth]{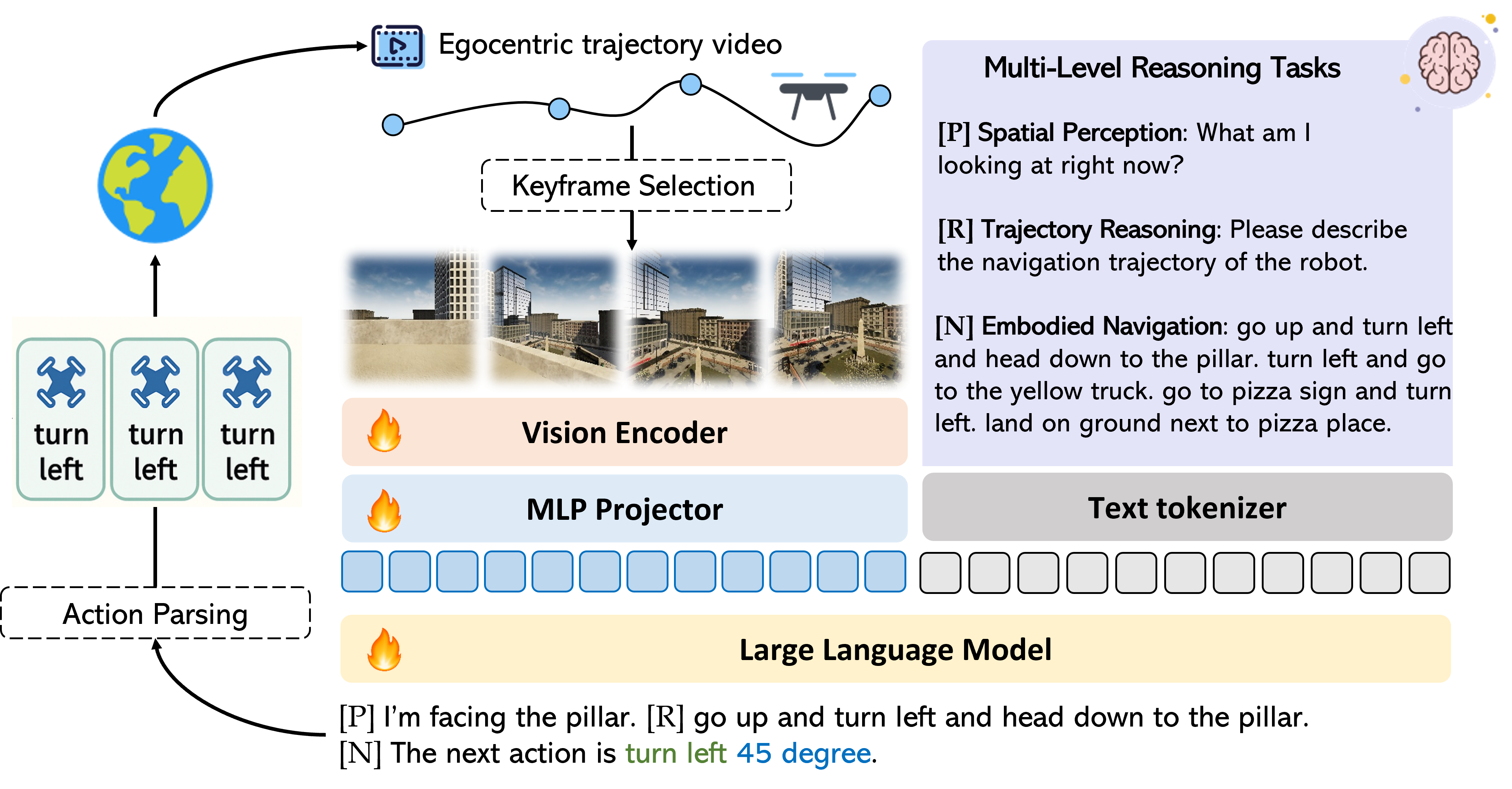}
    \caption{ Overview of our framework.
Given egocentric keyframes selected from the onboard video stream, our model first encodes the visual observations through a vision encoder and an MLP projector to obtain visual tokens, while language instructions are processed by a text tokenizer. The unified multimodal tokens are then fed into a large language model that is jointly trained on three complementary tasks: (i) Spatial Perception, which queries the current scene; (ii) Trajectory Reasoning, which summarizes historical motion and infers the agent’s navigational context; and (iii) Embodied Navigation, which predicts high-level action commands. The predicted textual action is further parsed and decomposed into a sequence of predefined motion primitives for execution in the physical environment.
    }
    \label{fig:model}
\end{figure*}

\subsection{Data Preprocessing} In data preprocessing stage, we refine the raw drone trajectories by applying action merging and keyframe selection, two complementary steps that improve the temporal structure of training samples. Together, these preprocessing steps organize trajectories into clearer motion segments paired with representative visual cues, improving the quality of supervision for navigation learning. As shown in Fig.~\ref{fig:preprocess_vis}, our approach expands the action 
vocabulary, produces a more balanced action distribution, and significantly 
shortens trajectories by eliminating redundant micro-actions.

\noindent \textbf{Action Merging.} UAV navigation trajectories are typically long and dominated by frequent forward micro-steps, leading to highly imbalanced and fragmented action sequences. To reduce fragmented micro-actions and enhance the semantic clarity of motion patterns, we merge consecutive identical actions into a single segment with a bounded length (e.g., combining three turn left $15^\circ$ steps into a single turn left $45^\circ$ step). This produces more meaningful action units and yields a more balanced and diverse action distribution. 

\noindent \textbf{Keyframe Selection.} To further suppress temporal visual redundancy, we extract keyframes at the boundaries of the merged action segments. This simple yet effective strategy is motivated by the observation that abrupt changes in the agent’s motion—such as transitioning from straight movement to a turning action—are often caused by the appearance of landmarks. These turning points naturally delimit semantically meaningful transitions in the underlying trajectory. By selecting the frames at these boundaries as keyframes, we obtain a compact set of observations that preserves landmarks and other salient visual cues while filtering out redundant intermediate views. This yields a more informative and temporally structured visual stream for downstream navigation learning. Unlike simple uniform sampling \cite{cheng2025navila} or detector-dependent landmark selection \cite{gao2025openfly}, our strategy is control-aligned and does not rely on auxiliary perception modules, providing an efficient keyframe selection mechanism.

\subsection{Framework Overview}

As illustrated in Fig. \ref{fig:model}, our framework tackles the aerial vision-and-language navigation task by processing natural-language instructions together with an egocentric video stream under a unified next-token prediction paradigm.

\noindent \textbf{Overall Pipeline.}  At each timestep, we sample a compact set of keyframes from the observation trajectory so far and pair them with the instruction. The keyframes are encoded into visual tokens through a vision encoder and projected into the language embedding space, while the instruction is tokenized independently. The resulting multimodal token sequence is fed into a large language model (LLM), which produces the next action directly in text form through autoregressive decoding. This action text is parsed into a predefined action set of low-level control commands and executed to update the drone’s state in 3D aerial environments.  To strengthen the agent’s spatial understanding and its ability to interpret navigational progress, we introduce two auxiliary tasks, termed spatial perception and trajectory reasoning. The spatial perception task guides the agent to answer egocentric questions about the current scene, whereas the trajectory reasoning task requires summarizing the observation trajectory. Together, these tasks provide complementary supervision that refines the model’s representation of spatial structure and navigation dynamics, ultimately boosting aerial navigation performance.

\begin{figure}[t!]
\color{black}
    \centering
    \includegraphics[width= \linewidth]{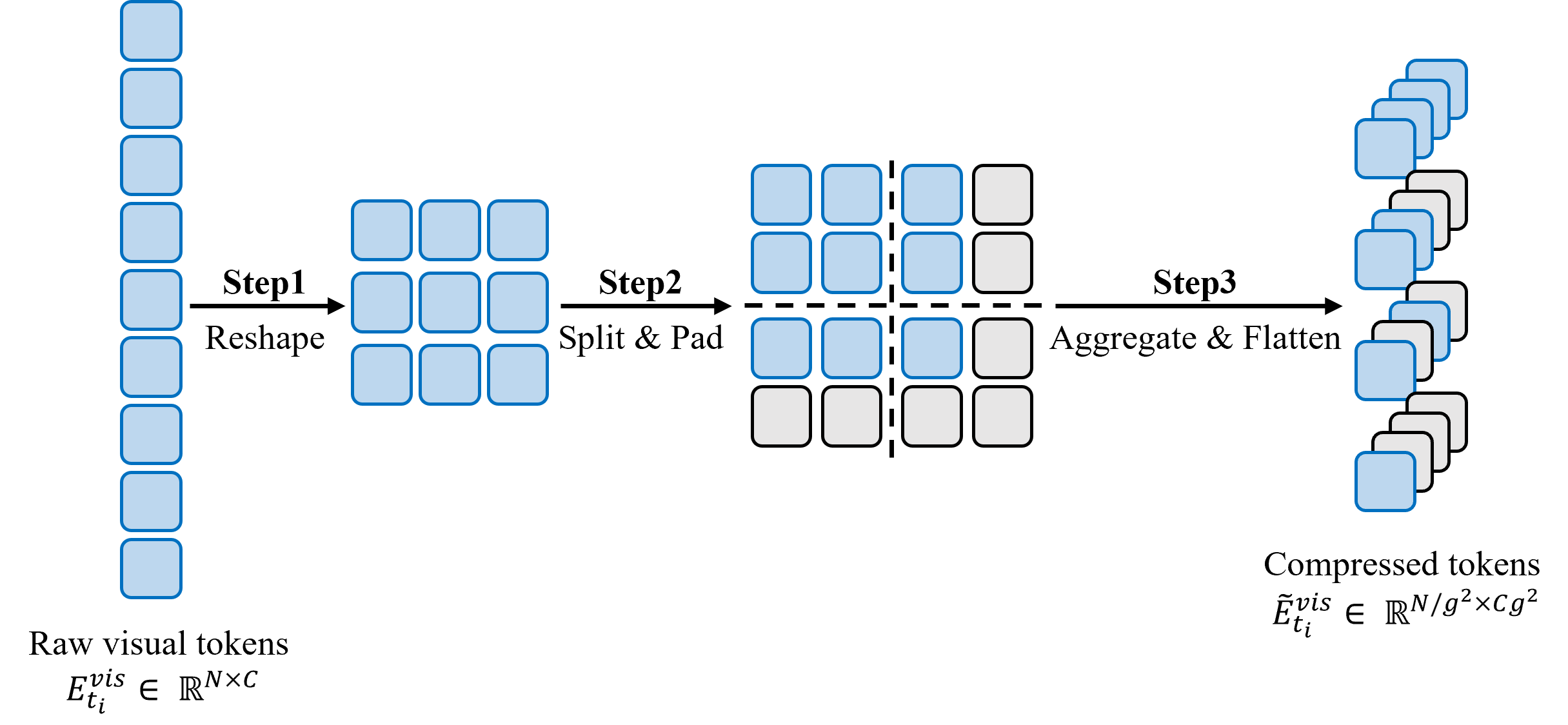}
    \caption{Schematic diagram of the STC module with grid size $g$ = 2.}
    \label{fig:stc}
\end{figure}

\noindent \textbf{Observation Encoding and Multimodal Tokenization.} At each timestep $t$, the agent receives the current egocentric image $x_t$
together with the accumulated visual trajectory. To obtain a compact yet
informative representation, we uniformly sample $K$ frames while always
retaining the first frame and the current frame. The first frame provides a
stable reference of the starting viewpoint, whereas the current frame serves
as the primary basis for predicting the next navigation action; the remaining
historical frames supply contextual cues for interpreting navigation progress. Let the sampled keyframes be 
$\mathcal{X}_t = \{x_{1}, x_{t_2}, \ldots, x_{t_K}\}$ 
with $x_{t_K} = x_t$. Each frame is encoded by a vision backbone into patch-level features:
\begin{equation}
    E^{\text{vis}}_{t_i} = \mathrm{Enc}_{\text{vis}}(x_{t_i}) 
    \in \mathbb{R}^{N \times C}.
\end{equation}

Long-horizon trajectories may yield prohibitively long visual token sequences.
To reduce the length of token sequences while preserving local spatial context, we introduce a Spatial Token Compression (STC) module, as illustrated in Fig. \ref{fig:stc}. Given patch-level visual tokens $E^{\text{vis}}_{t_i} $, STC first reshapes the token sequence into a 2D feature map. The feature map is then partitioned into non-overlapping 
$g \times g$ grids, where 
$g$ denotes the  grid size. When the spatial resolution is not divisible by 
$g$, zero-padding is applied along the corresponding dimensions before grid partitioning. Within each grid, the 
$g \times g$ tokens are concatenated along the channel dimension and flattened to form a compact visual representation. This yields a compressed token sequence:
\begin{equation}
\color{black}
\label{eq:stc}
    \tilde{E}^{\text{vis}}_{t_i} = \mathrm{STC}(E^{\text{vis}}_{t_i})
    \in \mathbb{R}^{(N/g^2)\times(Cg^2)}.
\end{equation} A lightweight MLP projector aligns visual features to the LLM embedding
space:
\begin{equation}
    Z^{\text{vis}}_{t_i}
        = \mathrm{Proj}_{\text{mlp}}(\tilde{E}^{\text{vis}}_{t_i})
           \in \mathbb{R}^{(N/g^2)\times D}.
\end{equation}

The navigation instruction 
$I = \{\omega_1,\ldots,\omega_L\}$ 
is tokenized by the LLM tokenizer to obtain text tokens
$Z^{\text{text}}\in\mathbb{R}^{L\times D}$.
We then concatenate visual and textual tokens into a unified multimodal
sequence:
\begin{equation}
\color{black}
    S_t = [Z^{\text{vis}}_{1},\, \ldots,\, Z^{\text{vis}}_{t_K},\, Z^{\text{text}}],
\end{equation} 
which serves as the observation representation for navigation.
The
sequence is fed into the LLM backbone, yielding the hidden state:
\begin{equation}
    H_t = \mathrm{LLM}(S_t),
\end{equation}
from which the next action $a_t$ is autoregressively generated in linguistic form.

\begin{figure}[t!]
    \centering
    \includegraphics[width=\linewidth]{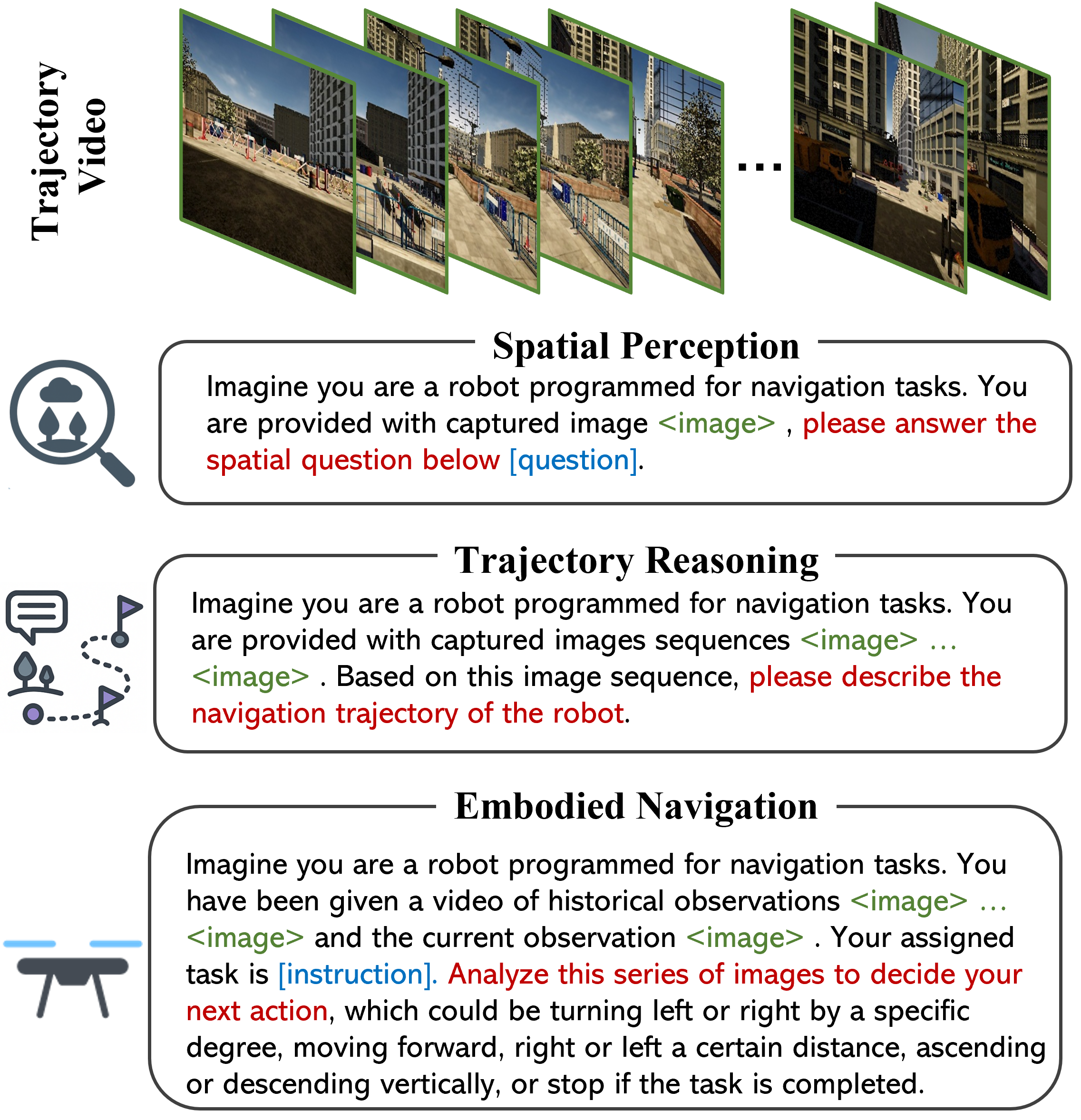}
    \caption{
    Unified prompting interface for the proposed model.  Through task-specific prompts, the model supports aerial navigation as the primary task, while also handling spatial perception and trajectory reasoning as auxiliary capabilities that enrich spatial understanding and temporal grounding.
    }
    \label{fig:prompt}
\end{figure}
\noindent \textbf{Unified Prompting Interface.} Aerial VLN poses substantially higher demands on scene understanding and
temporal reasoning due to the semantic complexity of outdoor 3D environments
and the long-horizon navigation trajectories. Effective navigation
requires the agent to interpret fine-grained spatial cues, maintain awareness
of its progression along the trajectory, and ground language instructions to
visual observations across a large spatial scale. To support the acquisition
of these complementary reasoning abilities, we adopt a unified prompting
interface that formulates all tasks within the same next-token
prediction paradigm, without any modification to the underlying model architecture. As shown in Fig.~\ref{fig:prompt}, we instantiate three prompt-driven task
formulations that expose different levels of navigation-relevant reasoning:
\textbf{Spatial Perception}, \textbf{Trajectory Reasoning}, and 
\textbf{Embodied Navigation}. Spatial perception prompts require the model to
answer egocentric, scene-centric questions based on the current view 
(e.g., ``What object appears on the right side of the image?''), thereby
reinforcing fine-grained geometric and semantic grounding. Trajectory
reasoning prompts ask the model to summarize the observed image sequence, encouraging a structured understanding of navigation progress.
Embodied navigation prompts instruct the model to predict the next control step 
in linguistic form. These task formulations provide complementary supervisory signals that
strengthen the model’s multi-level reasoning over spatial structure and
temporal trajectory evolution, thereby supporting more reliable navigation
decision-making.

\begin{figure}[t!]
\color{black}
    \centering
    \includegraphics[width= 0.95 \linewidth]{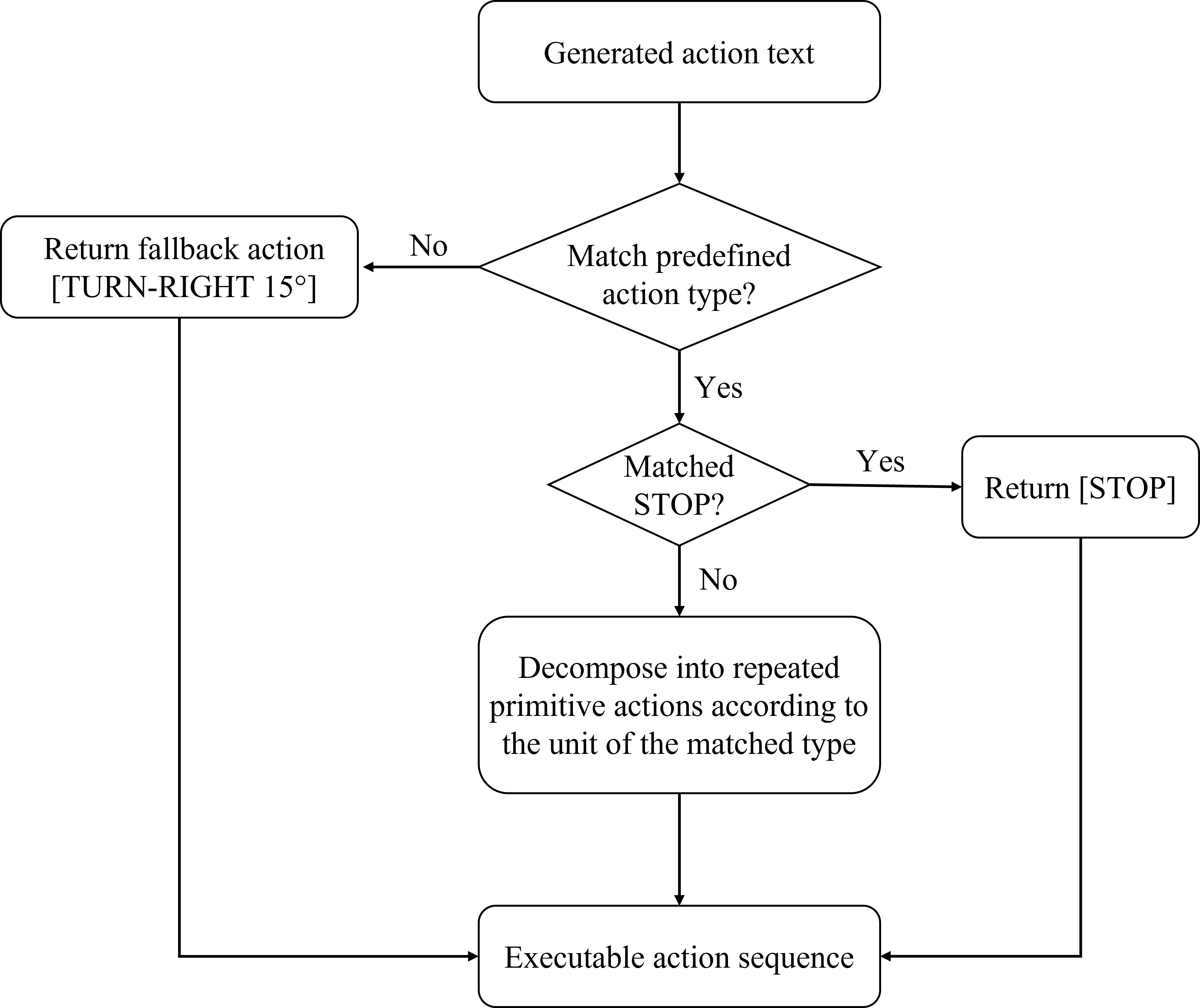}
    \caption{Workflow of the regular-expression action parser that converts generated textual actions into executable primitive action sequences.}
    \label{fig:action_parse}
\end{figure}

\noindent \textbf{Action Parsing and Execution.} During inference, the model operates with the embodied navigation-specific prompt as illustrated in Fig. \ref{fig:prompt}, and directly generates textual navigation actions. Once the model outputs a textual action command (e.g., “The next action is move forward 15 units”), we apply a regular-expression parser to extract the action type and its numeric argument. The parsed action is then decomposed into low-level motion primitives (e.g., multiple fixed-step forward movements). A schematic illustration of the overall parsing procedure is shown in Fig.~\ref{fig:action_parse}, with additional details provided in Appendix~\ref{app:action_parser}. These primitives are executed in an open-loop manner until all of them are completed, after which the agent receives the next visual observation from the environment, updates its trajectory history, and predicts the subsequent action. We quantitatively evaluated the reliability of this parsing module across simulated benchmarks, where it achieved a 100\% success rate in extracting valid actions. This design follows a robust and established paradigm in vision-language navigation research, with prior works such as NaVid \cite{zhang2024navid} and NaVid-4D \cite{Navid-4D} reporting equally reliable performance using the same approach in both simulation and real-world deployment.

\subsection{Training Objective}
Our model is trained in a multi-task setting that jointly supervises aerial
navigation, spatial perception, and trajectory reasoning through
autoregressive language modeling. Since navigation trajectories exhibit a
highly imbalanced distribution of actions, we apply a frequency-based
label reweighting strategy. Each action $a \in \mathcal{A}$ is assigned a
normalized inverse-frequency weight.
\begin{equation}
\label{eq:action_weight}
w(a)=
\sqrt{
    \frac{
        1/p(a)
    }{
        \frac{1}{|\mathcal{A}|}\sum_{b\in\mathcal{A}} 1/p(b)
    }
}.
\end{equation}

The overall multi-task objective integrates supervision from all three tasks.
While $\mathcal{B}$ denotes the mixed training set, the loss is defined as
\begin{equation}
\label{eq:final_loss}
\begin{aligned}
\mathcal{L}
    &= 
    \frac{1}{|\mathcal{B}|}
    \sum_{u\in\mathcal{B}}
    W(u)\,\ell_{\mathrm{CE}}(y_u,z_u), \\[6pt]
W(u)
    &= 
    \begin{cases}
        w(a_t),              & u\in\mathcal{D}_{\text{nav}}, \\[4pt]
        \lambda_{\text{sp}}, & u\in\mathcal{D}_{\text{sp}},  \\[4pt]
        \lambda_{\text{tr}}, & u\in\mathcal{D}_{\text{tr}},
    \end{cases} \\[6pt]
\mathcal{B}
    &=
    \mathcal{D}_{\text{nav}}
    \cup
    \mathcal{D}_{\text{sp}}
    \cup
    \mathcal{D}_{\text{tr}}.
\end{aligned}
\end{equation}
where $W(u)$ denotes the task-specific weight assigned to sample $u$. The term
$\ell_{\mathrm{CE}}(y_u, z_u)$ is the standard autoregressive
cross-entropy loss computed between ground truth $y_u$ and model
output $z_u$. For samples from other auxiliary tasks, we apply task-specific weighting to
regulate their relative influence during optimization. $\lambda_{\text{sp}}$ and $\lambda_{\text{tr}}$ are the weighting hyperparameters for
spatial perception and trajectory reasoning samples, respectively.
\begin{table*}[t]
\centering
\renewcommand{\arraystretch}{1.3}
\caption{Overview of the curated datasets for multi-task training}
\begin{tabular}{lccc}
\hline
\textbf{Task Type} &  \textbf{Data Source} & \textbf{Description} \\
\hline

Spatial Perception   &   Open3D-VQA (Sim env) + GQA (Real env) &  Drone egocentric spatial reasoning + spatial-relation  QA pairs   \\
Trajectory Reasoning  & AerialVLN (Sim env) & Sub-trajectory segmentation + progress summaries  \\
Aerial Navigation    & AerialVLN (Sim env) & Step-wise tuples $(\tau_t, I, a_t)$ after action merging and keyframe selection  \\
\hline
\end{tabular}
\label{tab:dataset_overview}
\end{table*}

\subsection{Implementation Details}
\noindent \textbf{Dataset Curation.} To support the multi-task training objectives, we curate a
mixed dataset that provides complementary supervision for action prediction,
spatial grounding, and temporal reasoning, as shown in Table \ref{tab:dataset_overview}. Our curation is guided by two
principles: maintaining a coherent input-output structure across tasks and
constructing a balanced mixture that captures the diversity of aerial
navigation scenarios. For the aerial navigation task, we derive stepwise supervision
from AerialVLN simulation trajectories by applying action merging and keyframe
selection, producing compact training tuples $(\tau_t, I, a_t)$. For spatial perception, we incorporate
VQA-style supervision from the drone-centric subset of Open3D-VQA~
\cite{zhang2025open3dvqa}, complemented by a spatial-relation-focused subset
of GQA~\cite{hudson2019gqa} filtered for geometric predicates. 

To provide temporal abstraction supervision, we manually construct trajectory-reasoning annotations by segmenting training trajectories into ordered sub-trajectories and associating each segment with a corresponding sub-instruction. The segmentation is guided by semantic coherence and explicit endpoint grounding, while each sub-trajectory is constrained to span approximately 4--16 keyframes after action merging to avoid excessive granularity. The sub-instructions are directly extracted from the original instruction without paraphrasing or language-model generation, thereby preserving the original linguistic distribution. After excluding severely ambiguous cases, we annotate approximately 3k training trajectories and further construct around 10k trajectory-reasoning training instances through compositional augmentation. Additional details of the annotation protocol are provided in Appendix~\ref{app:tr_annotation}.
This curated mixture forms a coherent and comprehensive data blend for supervised fine-tuning of the
proposed model.

\noindent \textbf{Model Training.}
We fine-tune NVILA-lite-2B and NVILA-lite-8B \cite{liu2025nvila} on our aerial navigation task, initializing both models from their pretrained weights. Following NVILA, the vision encoder is implemented using SigLIP \cite{zhai2023sigclip}, the projector is a lightweight two-layer MLP that aligns visual features with the language embedding space, and the language model is based on Qwen2 \cite{yang2024qwen2technicalreport}. Specifically, we adopt a full fine-tuning strategy, where all model components, including the vision encoder, MLP projector, and large language model, are jointly optimized, with all parameters being trainable during training. We perform supervised fine-tuning (SFT) for our tasks and the training objective is defined by the loss function presented in Eq. \ref{eq:final_loss}.
Training is performed on 4 NVIDIA A100 GPUs using DeepSpeed ZeRO-3 optimization, with a learning rate of $2\times10^{-5}$. We employ a cosine decay schedule with a warm-up ratio of 0.03. We adopt a global batch size of 128, achieved through a per-GPU batch size of 8 and 4 gradient accumulation steps. All models are trained for 1 epoch. For action merging, consecutive identical actions are combined, with the maximum merge length limited to 3 steps. For keyframe sampling, we uniformly sample $K = 8$ frames from the visual trajectory, while ensuring that the first and the current frames are always preserved. We set the auxiliary task weights to $\lambda_{\text{sp}} = 1.0$ for spatial perception and $\lambda_{\text{tr}} = 0.5$ for trajectory reasoning.

\section{Experiments}
\subsection{Experimental Setup}
\noindent \textbf{Simulated environments.}
We conduct experiments across multiple aerial VLN benchmarks for comprehensive evaluation. AerialVLN-S benchmark \cite{liu2023aerialvln} provides a realistic aerial navigation setting rendered in AirSim and Unreal Engine~4. 
The dataset covers 17 compact outdoor scenes representing diverse urban environments such as residential areas, industrial zones, parks, and villages. For evaluation, the dataset allocates 12 environments to the training and validation-seen splits, while the validation-unseen split is drawn from 5 distinct scenes.  Flight trajectories are collected by expert pilots, who follow naturalistic aerial routes to ensure realistic motion patterns and diverse path configurations. Overall, AerialVLN-S provides 10,113 language instructions for training, with 333 and 531 instructions assigned to the validation-seen and validation-unseen splits, respectively. 

In addition to AerialVLN-S, we evaluate our method on the OpenFly benchmark \cite{gao2025openfly}, constructed via automatic trajectory generation and vision-language-model-based instruction synthesis. OpenFly features diverse urban environments with varied scene layouts, flight heights, and trajectory lengths. Given the large scale of OpenFly, which contains over 100k navigation trajectories, we conduct experiments on a selected subset for practical evaluation. Specifically, we use trajectories collected from six high-quality UE4 scenes, forming a subset referred to as OpenFly-S. Training and evaluation are conducted on subsets comprising 61,004 and 1,210 trajectories, respectively. Detailed subset construction and per-scene statistics are provided in Appendix~\ref{app:openflys}.

\noindent \textbf{Evaluation Metrics.}
We adopt several standard metrics to evaluate aerial vision-and-language navigation\cite{liu2023aerialvln}:

\begin{itemize}[leftmargin=*]
    \item \textbf{Navigation Error (NE/$m$).} Measuring goal-reaching accuracy by computing the Euclidean distance between the UAV’s final position and the destination.

    \item \textbf{Success Rate (SR/$\%$).} Evaluating task completion as the proportion of episodes in which the UAV stops within a 20-meter radius of the destination.

    \item \textbf{Oracle Success Rate (OSR/$\%$).} Relaxing the success criterion by marking an episode successful if \emph{any} point along the predicted trajectory falls within this 20-meter radius.

    \item \textbf{Success-weighted Dynamic Time Warping (SDTW/$\%$).} Assessing trajectory alignment by applying normalized DTW between the predicted and ground-truth paths and weighting the score by the Success Rate.

    \item \textbf{Success weighted by Path Length (SPL/$\%$).} Assessing navigation efficiency by weighting task success with path optimality, computed as the ratio between the shortest path length and the executed path length for successful trajectories.
    
\end{itemize}

\noindent \textbf{Baselines.} We compare our approach against previous diverse baselines. These baselines fall into three categories  \cite{zhang2025citynavagent}:

\begin{itemize}[leftmargin=*]
    \item \textbf{Statistical-based methods.} 
    Random Sampling uniformly samples actions from the action space while 
Action Sampling selects actions according to the training-set action distribution.

    \item \textbf{Zero-shot LLM-based methods.} 
MapGPT~\cite{chen2024mapgpt} introduces a language-based online map representation to equip GPT-4v \cite{achiam2023gpt} with global spatial understanding, enabling adaptive multi-step planning for indoor VLN in a zero-shot setting.
STMR\cite{gao2024aerialstmr} enhances spatial reasoning by introducing a Semantic-Topo-Metric Representation that allows LLMs to infer navigation decisions directly from structured scene information. 
CityNavAgent\cite{zhang2025citynavagent} performs hierarchical semantic planning to predict waypoints in a zero-shot manner, supported by a global memory module that records past trajectories for improved long-horizon navigation. 

    \item \textbf{Learning-based methods.} 
    LingUNet \cite{misra2018mapping} is an instruction-conditioned visual model adapted from goal-visible aerial navigation to step-wise VLN in AerialVLN\cite{liu2023aerialvln}.
    Seq2Seq \cite{anderson2018seq2se} predicts actions using a recurrent policy that encodes visual observations and language instructions into a unified hidden representation. CMA \cite{krantz2020cma} introduces cross-modal attention to better fuse RGB, depth, and textual cues for action prediction. LAG\cite{liu2023aerialvln} adds a look-ahead guidance module. Grid-VS\cite{zhao2025aerialviewselection} formulates aerial VLN as a grid-based view selection task and leverages a bird’s-eye-view representation with a cross-modal transformer to model long-term navigation context and instruction alignment.
    NaVid \cite{zhang2024navid} is a video-based vision-language model that has demonstrated strong performance in indoor VLN.  OpenFly \cite{gao2025openfly} adopts a keyframe-aware VLN architecture and uses adaptive token sampling over historical observations.
\end{itemize}

\begin{table*}[t]
\centering
\renewcommand{\arraystretch}{1.3}
\caption{Comparison of different methods on the AerialVLN-S dataset. Observations include single RGB (S.RGB), depth (Depth), panoramic view (Pano.), and odometry information (Odo.)}
\label{tab:results_airvln_s}
\resizebox{\textwidth}{!}{
\begin{tabular}{lcccc|cccc|cccc}
\hline
\textbf{Method} &
\multicolumn{4}{c}{\textbf{Observation}} &
\multicolumn{4}{c}{\textbf{Validation Seen}} &
\multicolumn{4}{c}{\textbf{Validation Unseen}} \\ 
\cline{2-5} \cline{6-9} \cline{10-13}
 & S.RGB & Depth & Pano. & Odo.
 & NE$\downarrow$ & SR$\uparrow$ & OSR$\uparrow$ & SDTW$\uparrow$
 & NE$\downarrow$ & SR$\uparrow$ & OSR$\uparrow$ & SDTW$\uparrow$ \\
\hline

Grid-based VS \cite{zhao2025aerialviewselection}&  &  & $\checkmark$ & $\checkmark$
 & 70.3 & 20.8 & 33.4 & 10.2
 & 121.3 & 7.4 & 16.1 & 2.5 \\

CityNavAgent \cite{zhang2025citynavagent} &  & $\checkmark$ & $\checkmark$  & $\checkmark$
 & 80.8 & 13.9 & 30.2 & 5.1
 & 60.2 & 11.7 & 35.2 & 5.0 \\

\hline

Random Sampling &  &  &  & 
 & 109.6 & 0.0 & 0.0 & 0.0
 & 149.7 & 0.0 & 0.0 & 0.0 \\

Action Sampling &  &  &  &
 & 213.8 & 0.9 & 5.7 & 0.3
 & 237.6 & 0.2 & 1.1 & 0.1 \\

LingUNet \cite{misra2018mapping} & $\checkmark$ & $\checkmark$ &  &
 & 383.8 & 0.6 & 6.9 & 0.2
 & 368.4 & 0.4 & 3.6 & 0.9 \\

Seq2Seq \cite{anderson2018seq2se} & $\checkmark$ & $\checkmark$ &  &
 & 146.0 & 4.8 & 19.8 & 1.6
 & 218.9 & 2.3 & 11.7 & 0.7 \\
 
CMA \cite{krantz2020cma}& $\checkmark$ & $\checkmark$ &  &
 & 121.0 & 3.0 & 23.2 & 0.6
 & 172.1 & 3.2 & 16.0 & 1.1 \\
 
LAG \cite{liu2023aerialvln} & $\checkmark$ & $\checkmark$ &  &
 & \underline{90.2} & 7.2 & 15.7 & 2.4
 & 127.9 & 5.1 & 10.5 & 1.4 \\

STMR \cite{gao2024aerialstmr}& $\checkmark$ & $\checkmark$ &  &
 & 96.3 & \textbf{12.6} & \underline{31.6} & \underline{2.7}
 & 119.5 & \textbf{10.8} & \underline{23.0} & \underline{1.9} \\

MapGPT  \cite{chen2024mapgpt}& $\checkmark$ &  &  &
 & 124.9 & 2.1 & 4.7 & 0.8
 & 107.0 & 0.0 & 0.0 & 0.0 \\

NaVid \cite{zhang2024navid} & $\checkmark$ &  &  &
 & 105.1 & 6.8 & 15.5 & 1.1
 & \underline{106.9} & 6.1 & 12.3 & 0.7 \\

OpenFly   \cite{gao2025openfly}& $\checkmark$ &  &  &
 & 127.2 & 8.1 & 21.8 & 1.6
 & 113.8 & 7.6 & 18.2 & 1.5 \\



  \textbf{Ours} & $\checkmark$ &  &  &
 & \textbf{79.6} & \underline{11.4} & \textbf{37.7} & \textbf{6.3}
 & \textbf{95.8} & \underline{8.1} & \textbf{28.9} & \textbf{2.2} \\

\hline
\end{tabular}}
\end{table*}

\begin{figure}[t!]
\color{black}
    \centering
    \includegraphics[width=\linewidth]{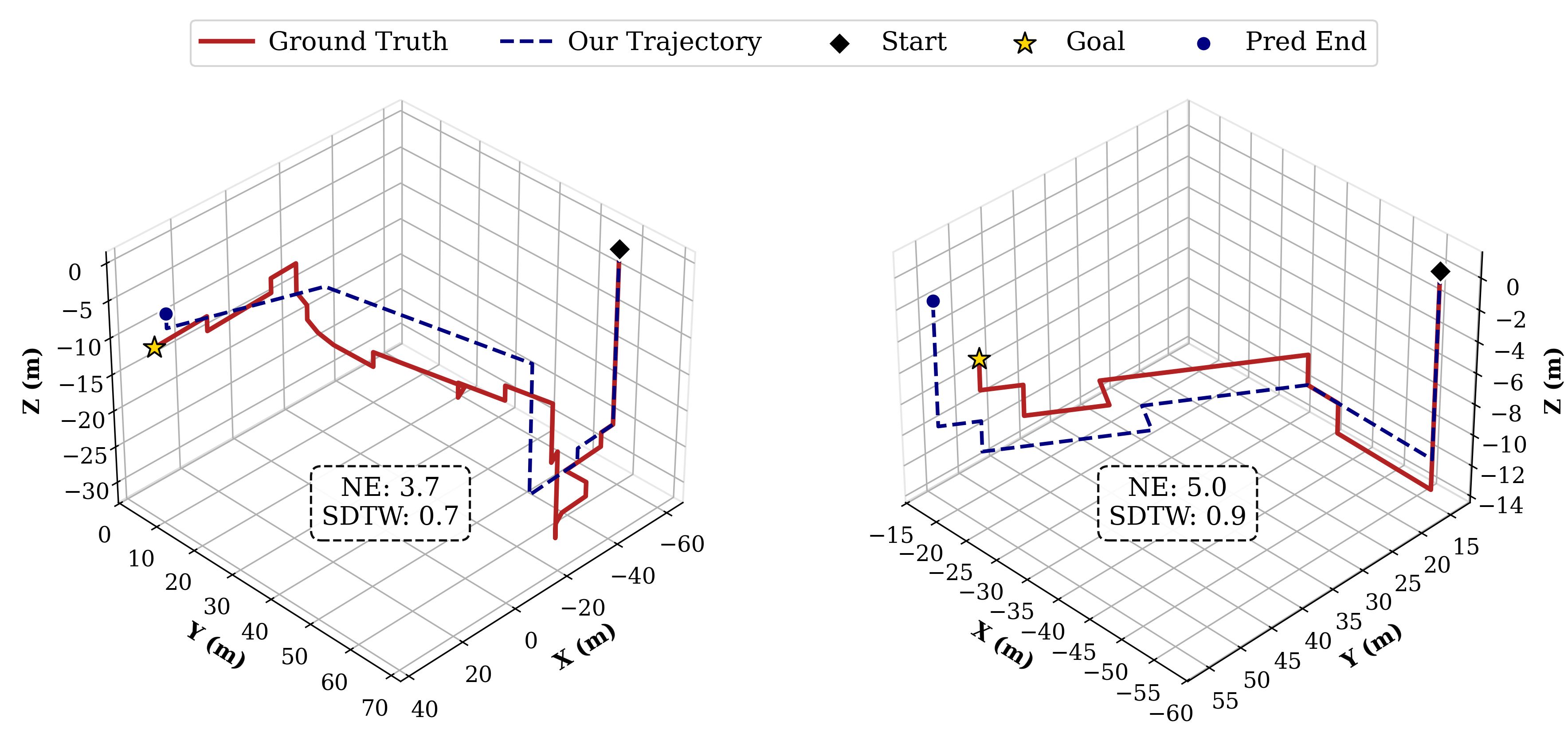}
    \caption{
    Qualitative comparison between predicted and ground-truth drone trajectories in 3D space.
Across multiple validation episodes, the predicted paths follow the global route structure, capturing major turns, long-range transitions, and altitude changes required by language instruction. These results highlight strong 3D spatial reasoning and stable flight control
from egocentric observations.
    }
    \label{fig:qualitative_3dtraj}
\end{figure}

\subsection{Results on Benchmarks}

\subsubsection{Quantitative Results on AerialVLN-S Benchmark}
To ensure a fair comparison, we categorize methods by their sensor configurations, distinguishing RGB-only agents from those that rely on panoramic inputs, depth sensing, or odometry in Table~\ref{tab:results_airvln_s}. Within the RGB-only group, our method achieves strong performance on all metrics and attains the best overall results, comparable to STMR \cite{gao2024aerialstmr}, which required external depth information. The proposed method is also competitive with panoramic baselines, especially in SDTW. This is particularly meaningful in the AerialVLN setting, where many trajectories contain turn-backs and repeated segments, making SDTW a more informative measure of instruction following than SR alone. Panoramic methods typically require multiple cameras or repeated in-place rotations to build 360° observations, which increases time and energy costs for UAVs, whereas our model operates from a single-view RGB stream and remains competitive under a more constrained monocular setting. MapGPT \cite{chen2024mapgpt} and NaVid \cite{zhang2024navid}, which are strong indoor VLN baselines, perform notably worse in outdoor aerial settings, underscoring the additional challenges posed by large-scale, sparsely structured environments. Compared with the most related OpenFly \cite{gao2025openfly} framework, our approach further benefits from long-horizon history modeling and auxiliary tasks, leading to more stable long-range behavior. Overall, these results establish a new state-of-the-art for monocular RGB-only aerial VLN  and validate the effectiveness of our unified framework for spatial, temporal, and embodied reasoning. Additional multi-seed statistics are provided in Appendix~\ref{app:multiseed}. Fig.~\ref{fig:qualitative_3dtraj} compares the predicted 3D flight paths with the 
reference routes. The predicted trajectories preserve the 
overall geometric structure of the routes, including the major turning segments,
long-range direction changes, and the characteristic height variations implied 
by the instruction. Although small local deviations occasionally appear, the 
agent often recovers from these offsets and continues to produce geometrically 
coherent paths over extended horizons. The trajectories also converge toward the 
intended final region, reflecting stable motion patterns derived solely from 
egocentric RGB observations in large-scale outdoor environments.

\begin{table}[t]
\color{black}
\centering
\renewcommand{\arraystretch}{1.3}
\caption{Comparison of different methods on the OpenFly-S test-seen split}
\label{tab:results_openfly_s}
\resizebox{\linewidth}{!}{
\begin{tabular}{lcc|cccc}
\hline
\textbf{Method} &
\multicolumn{2}{c}{\textbf{Observation}} &
\multicolumn{4}{c}{\textbf{OpenFly-S Test Seen}} \\ 
\cline{2-3} \cline{4-7} 
 & S.RGB & Depth
 & NE$\downarrow$ & SR$\uparrow$ & OSR$\uparrow$ & SPL$\uparrow$ \\
\hline
Random Sampling  & & & 242 & 0.7 & 0.8 & 0.0 \\
Seq2Seq \cite{anderson2018seq2se} & $\checkmark$ & $\checkmark$
 & 291 & 0.1 & 3.2 & 0.6 \\
CMA \cite{krantz2020cma} & $\checkmark$ & $\checkmark$
  &  144 & 5.7 & 30.0 & 5.1 \\
  LAG \cite{liu2023aerialvln}  & $\checkmark$ & $\checkmark$
 &130 & 6.6 & 40.8 & 7.0 \\
NaVid \cite{zhang2024navid}  & $\checkmark$ & 
 &142 & 9.9 & 24.3 & 3.9 \\
OpenFly \cite{gao2025openfly}  & $\checkmark$ & 
 & \underline{98}& \underline{33.2} & \underline{63.5} & \underline{22.8} \\
Ours  & $\checkmark$ & 
 & \textbf{38}& \textbf{54.5 } & \textbf{72.1} & \textbf{49.9}    \\

\hline
\end{tabular}}
\end{table}

\begin{figure}[t!]
\color{black}
    \centering
    \includegraphics[width= 0.98 \linewidth]{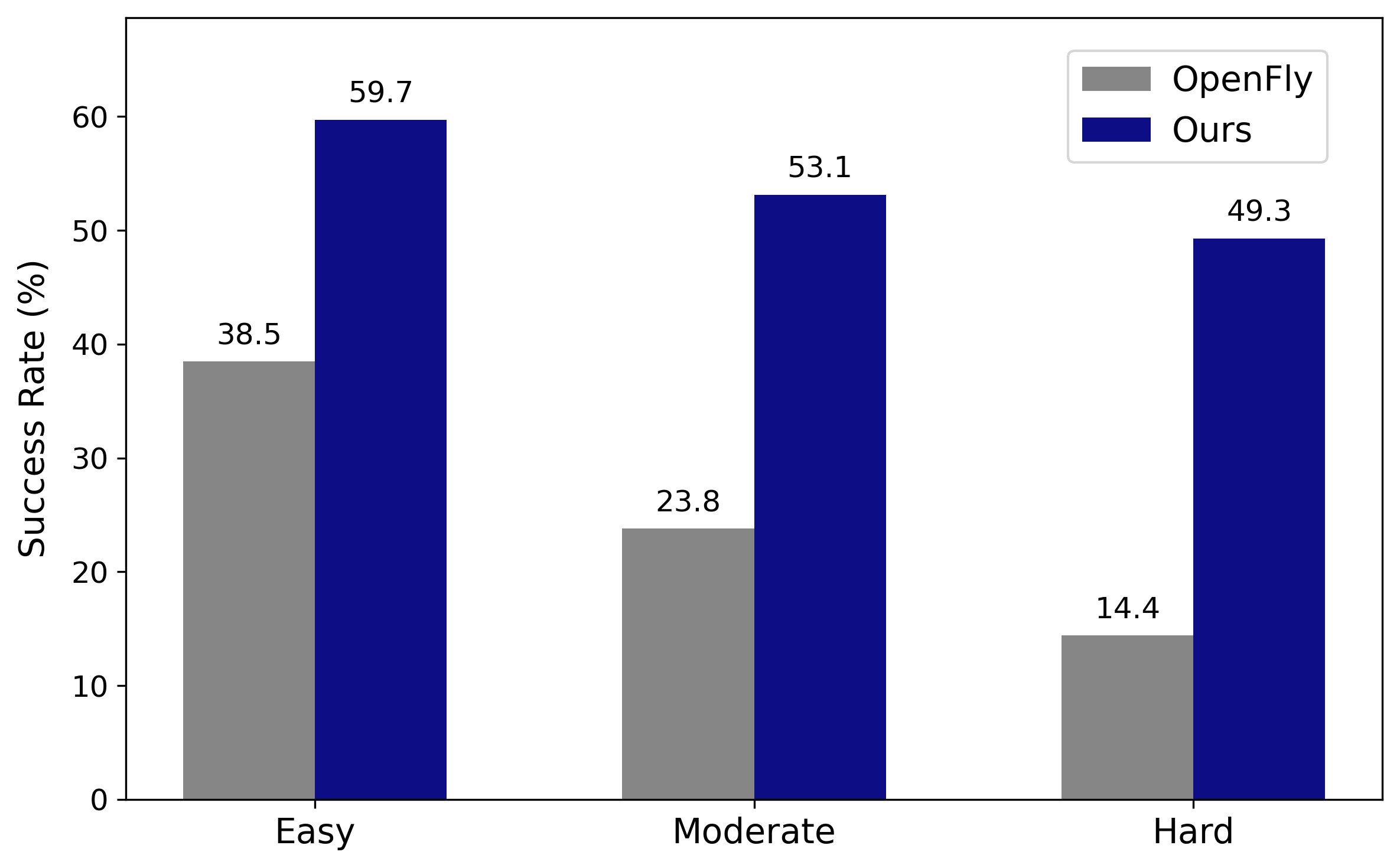}
    \caption{Comparison between OpenFly and our method across three difficulty levels (\textit{Easy}, \textit{Moderate}, and \textit{Hard}), grouped by trajectory length.}
    \label{fig:openfly-s}
\end{figure}

\subsubsection{Quantitative Results on OpenFly-S Benchmark} We report quantitative results on the OpenFly-S benchmark in Table~\ref{tab:results_openfly_s}. Our method achieves superior performance compared to baselines across all evaluation metrics, including NE, SR, OSR, and SPL. 
We further analyze success rates across different difficulty levels defined by trajectory length, following the evaluation protocol adopted in OpenFly \cite{gao2025openfly}. Specifically, episodes with fewer than 30 actions are categorized as \textit{Easy}, those with 30-60 actions as \textit{Moderate}, and those with more than 60 actions as \textit{Hard}. As shown in Fig. \ref{fig:openfly-s}, the proposed method beats OpenFly\cite{gao2025openfly} across all difficulty levels, with a larger margin on hard episodes. Notably, performance drops for both methods as the trajectory length increases, suggesting that long-horizon aerial navigation remains challenging. We observe that overall success rates on OpenFly-S are higher than those on AerialVLN-S. This behavior is consistent with the characteristics of the OpenFly dataset, which features shorter navigation trajectories and fewer action steps on average. Overall, the experimental results on both AerialVLN-S and OpenFly-S demonstrate that the proposed method maintains strong and consistent performance across multiple aerial VLN datasets constructed under different design principles, supporting its robustness to variations in trajectory scale, instruction length, and data generation strategies.

\begin{figure*}[t!]
    \centering
    \includegraphics[width=\textwidth]{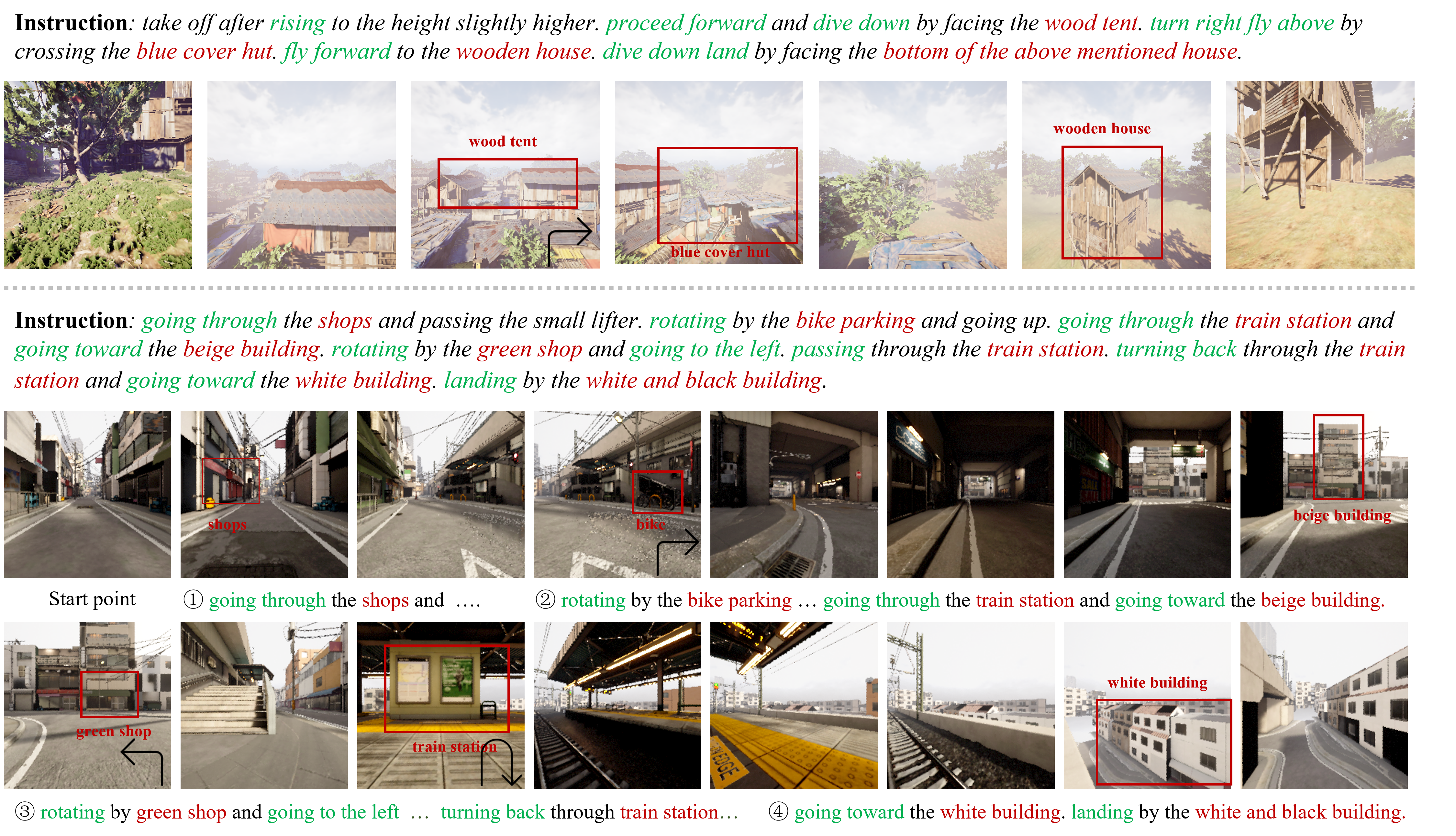}
    \caption{
    Visualizations of our method on AerialVLN-S benchmark.  
    Our model successfully follows detailed long-horizon instructions, grounds visual 
    landmarks, and executes correct spatial maneuvers across complex outdoor scenes, 
    demonstrating strong instruction-following consistency and robust navigation ability. For clarity, colored elements denote \textcolor{red}{landmarks} and \textcolor{green}{action phrases}. Trajectories are visually grouped to facilitate analysis of long-horizon behaviors.
    }
    \label{fig:qualitative}
\end{figure*}

\subsubsection{Cross-dataset Evaluation}
Cross-dataset generalization is a critical capability for vision-and-language navigation models. Inspired by the evaluation protocol in ground VLN \cite{zhang2024navid,cheng2025navila,wang2025monodream}, we assess cross-dataset generalization in aerial VLN by training models on AerialVLN-S and directly evaluating them on AerialVLN without any fine-tuning. Compared to AerialVLN-S, AerialVLN presents a more challenging navigation scenario, with longer path lengths, larger-scale environments, and more complex long-horizon action sequences \cite{liu2023aerialvln}. As shown in Table~\ref{tab:results_cross_dataset}, all methods exhibit clear performance degradation under the cross-dataset setting, reflecting the substantial shift from AerialVLN-S to AerialVLN. Nevertheless, our method remains superior to Seq2Seq \cite{anderson2018seq2se} and CMA \cite{krantz2020cma} on all reported metrics, indicating stronger transferability. To further evaluate the effect of auxiliary supervision, we compare our method with the variant trained without auxiliary tasks. The improvements in SR, OSR, and SDTW suggest that auxiliary tasks enhance the agent’s spatial and temporal reasoning over the environment, thereby improving robustness under distributional shift.

\begin{table}[t]
\color{black}
\centering
\renewcommand{\arraystretch}{1.3}
\caption{Cross-dataset Generalization Evaluation on AerialVLN Val-Unseen split. Note that $\dagger$ indicates our method without auxiliary tasks} %
\label{tab:results_cross_dataset}
\resizebox{\linewidth}{!}{
\begin{tabular}{lcc|cccc}
\hline
\textbf{Method} &
\multicolumn{2}{c}{\textbf{Observation}} &
\multicolumn{4}{c}{\textbf{AerialVLN Validation Unseen}} \\ 
\cline{2-3} \cline{4-7} 
 & S.RGB & Depth
 & NE$\downarrow$ & SR$\uparrow$ & OSR$\uparrow$ & SDTW$\uparrow$ \\
\hline

Seq2Seq \cite{anderson2018seq2se} & $\checkmark$ & $\checkmark$
 & 352.7 & 0.6 & 4.2 & 0.3 \\
CMA \cite{krantz2020cma} & $\checkmark$ & $\checkmark$
  & 337.0 & 1.0 & 6.7 & 0.4 \\
Ours$^\dagger$ & $\checkmark$  &  &\textbf{311.9}
 &1.8 & 12.1 &0.5  \\

Ours  & $\checkmark$ & 
 &331.5 &\textbf{2.8} & \textbf{13.3} & \textbf{0.8} \\

\hline
\end{tabular}}
\end{table}

\begin{figure*}[t!]
    \centering
    \includegraphics[width=\textwidth]{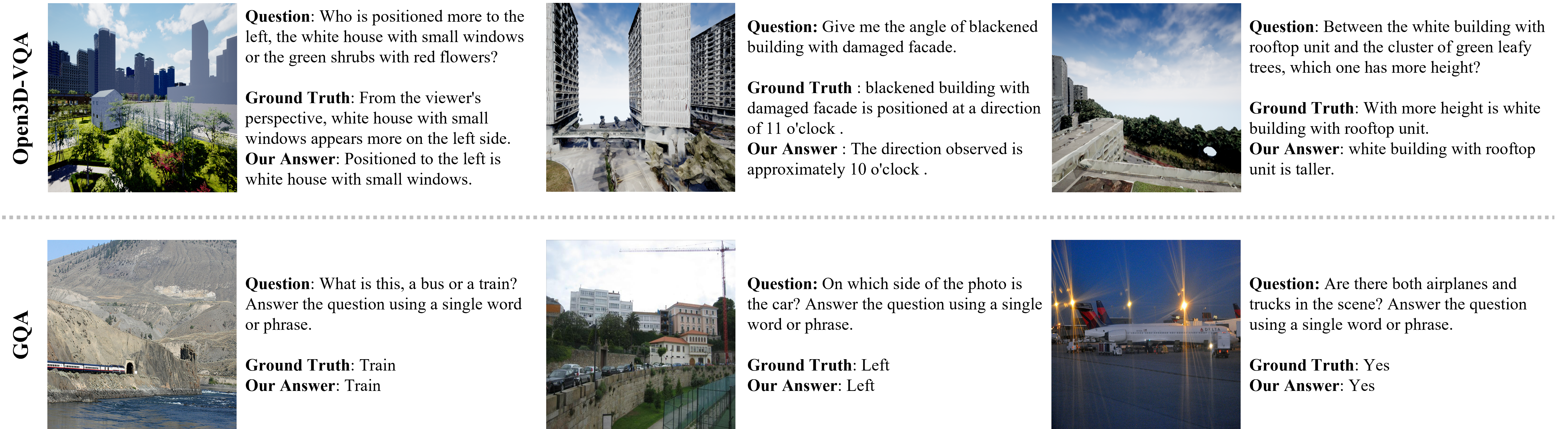}
    \caption{
    Qualitative results on auxiliary VQA tasks. Top: examples from the Open3D-VQA dataset \cite{zhang2025open3dvqa}, where the model answers 3D spatial queries 
including relative relations and geometric orientation. 
Bottom: examples from GQA \cite{hudson2019gqa}, covering object grounding 
and relational reasoning. These examples illustrate the model’s spatial 
understanding and positional reasoning, which is crucial for language-guided UAV navigation.
    }
    \label{fig:qualitative_vqa}
\end{figure*}

\begin{figure*}[t!]
    \centering
    \includegraphics[width=\textwidth]{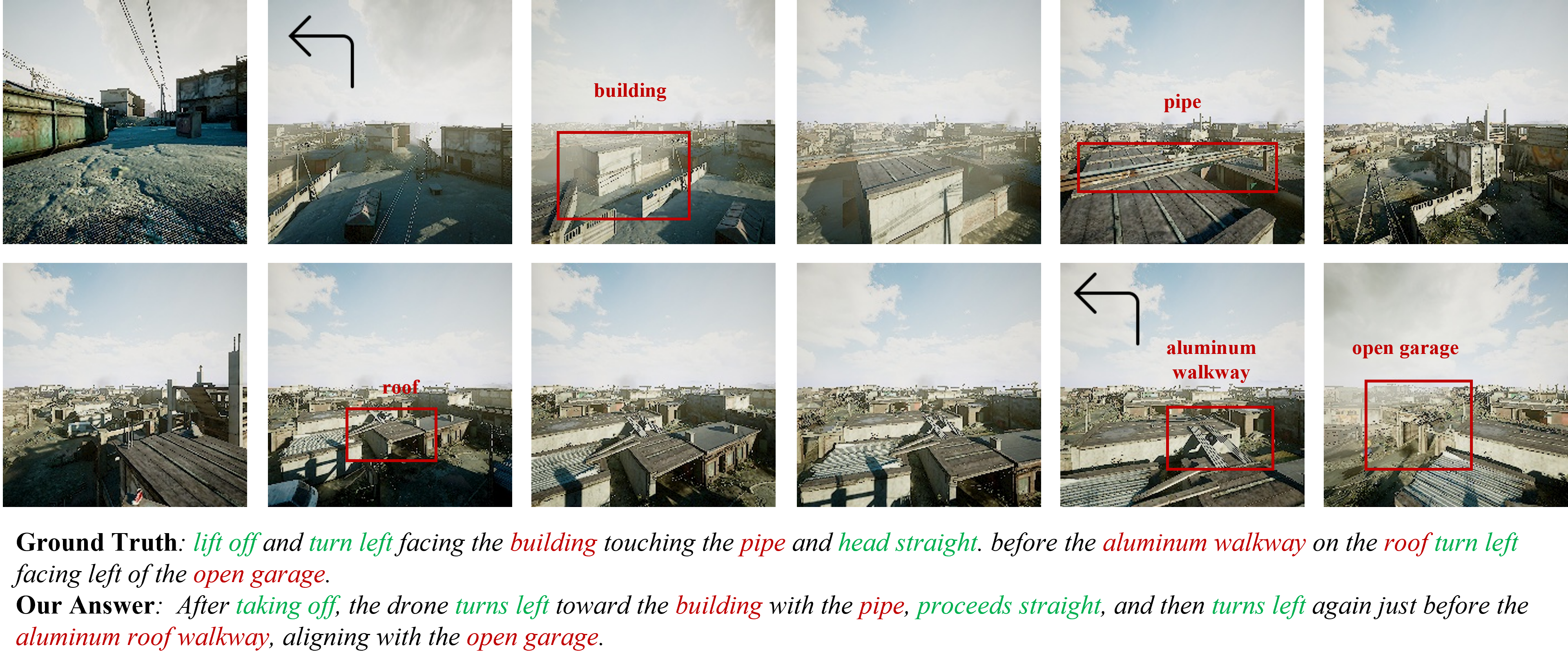}
    \caption{
    Qualitative results on trajectory summary task. The model summarizes the route by identifying key landmarks (building, pipe, walkway, garage) and the major motion segments (take-off, left turns, forward movement), demonstrating temporal understanding of the visual trajectory.
    }
    \label{fig:qualitative_summary}
\end{figure*}

\subsubsection{Qualitative Analysis} 
We provide qualitative navigation trajectories in Fig.~\ref{fig:qualitative} 
to assess the model's ability to interpret and ground natural language 
instructions in realistic outdoor environments. The first example illustrates a 
medium-length instruction including several landmark-based references. 
Our agent demonstrates reliable instruction-following behavior by consistently 
grounding entities such as \textit{shops}, \textit{bike parking}, and 
\textit{green shop}, and by executing the corresponding spatial maneuvers 
(e.g., \textit{turning, approaching, and passing}) at the appropriate visual cues. This indicates that the model preserves strong cross-modal alignment between linguistic expressions and visual observations, a fundamental capability for step-by-step decision-making in VLN. The second example showcases navigation under a significantly longer and more descriptive instruction spanning over one hundred actions. 
Throughout this extended trajectory, the agent maintains instruction 
consistency, preserves stable temporal grounding, and generates coherent 
movement decisions despite the increased complexity. These results highlight the robustness of our model in long-horizon instruction following, an important capability for aerial VLN in unconstrained outdoor environments.
Fig.~\ref{fig:qualitative_vqa} presents qualitative examples from the auxiliary VQA datasets used in training to illustrate the auxiliary reasoning capabilities of the model. On the Open3D-VQA dataset, the model correctly answers 3D spatial queries such as which object is higher or what lies to the left/right in an egocentric aerial view. On the GQA subset, it identifies objects and resolves simple spatial relations in visually cluttered scenes. These examples show that the model can interpret scene geometry and spatial layout from egocentric inputs.
Fig.~\ref{fig:qualitative_summary} presents qualitative examples from the 
trajectory reasoning task. Given the historical egocentric visual trajectory, the model 
produces a concise natural-language description that captures the high-level 
motion pattern and the principal scene transitions along the route. The generated summaries demonstrate consistent temporal abstraction by identifying turning segments, periods of forward movement, and transitions associated with spatial landmarks such as parks, intersections, and surrounding buildings. Moreover, the model reflects the causal structure of the trajectory by linking sequences of 
actions to the corresponding visual and semantic changes in the environment. These examples show that the model can integrate long-term dependencies and organize extended action sequences into coherent semantic units, which provides a complementary form of temporal grounding beyond low-level action prediction.

\begin{table}[t]
\centering
\color{black}
\renewcommand{\arraystretch}{1.3}
\caption{Comparison of Model Efficiency}
\label{tab:model_efficiency}
\begin{tabular*}{0.95\linewidth}{@{\extracolsep{\fill}} lcc}
\hline
Method  & Peak GPU Memory & Inference Latency \\
\hline
OpenFly (7B)   & $\sim$ 14 GB & 0.5 s \\
Ours (8B)  & $\sim$ 17 GB & 0.7 s \\
Ours (8B)-AWQ      & $\sim$ 6  GB  & 0.5 s \\

\hline
\end{tabular*}
\end{table}
\subsection{Model Efficiency}
We evaluate the inference efficiency of our method to examine its computational cost for aerial vision-language navigation. We compare our approach with OpenFly \cite{gao2025openfly}, a recent state-of-the-art aerial VLN method, as both methods adopt a VLM-based design paradigm. All measurements are conducted on a single NVIDIA A100 GPU. We report per-step end-to-end inference latency  \footnote{End-to-end inference latency is measured as the wall-clock time from receiving multimodal observations (RGB images and language instruction) to producing the predicted action.} together with peak GPU memory. As shown in Table \ref{tab:model_efficiency}, the full-precision 8B model exhibits a per-step inference latency of about 0.7 s and a peak GPU memory usage of approximately 17 GB. We further note that the favorable inference latency of OpenFly \cite{gao2025openfly} mainly benefits from its limited visual input, leading to lower multimodal processing overhead and reduced temporal visual context. To alleviate the memory overhead, we further evaluate an Activation-aware Weight Quantization (AWQ) \cite{lin2024awq} version of the 8B model. AWQ substantially reduces the memory footprint and lowers the inference latency to around 0.5 s, resulting in a decision rate close to 2 Hz. Further improvements in inference efficiency are expected to benefit from more advanced acceleration techniques and hardware-aware optimizations.

\subsection{Ablation Study}
\begin{table}[t]
\renewcommand{\arraystretch}{1.3}
\centering
\caption{Ablation study on auxiliary tasks: Spatial Perception (SP) and Trajectory Reasoning (TR). Both tasks improve success rate (SR) and trajectory quality (SDTW) in both seen and unseen environments}
\begin{tabular*}{0.95\linewidth}{@{\extracolsep{\fill}} c c | c c | c c}
\hline
\multicolumn{2}{c|}{\textbf{Auxiliary Tasks}} & \multicolumn{2}{c|}{\textbf{Val Seen}} & \multicolumn{2}{c}{\textbf{Val Unseen}} \\
SP & TR & SR$\uparrow$ & SDTW$\uparrow$ & SR$\uparrow$ & SDTW$\uparrow$ \\
\hline
           &             & 9.6  & 4.5 & 5.8  & 1.6 \\
\checkmark &             & 10.8 & 5.4 & 7.0  & 1.7 \\
\checkmark & \checkmark  & \textbf{11.4} & \textbf{6.3} & \textbf{8.1} & \textbf{2.2} \\
\hline
\end{tabular*}
\label{tab:auxiliary_tasks}
\end{table}

\subsubsection{Effect of Auxiliary Tasks} 
We conduct ablation experiments to evaluate the effect of the two auxiliary tasks: Spatial Perception (SP) and Trajectory Reasoning (TR) in Table~\ref{tab:auxiliary_tasks}. Without auxiliary supervision, the policy shows limited spatial grounding and produces the lowest SR and SDTW on both splits. SP leads to consistent gains, confirming that geometric cues improve local action grounding. Incorporating TR further boosts performance, with the largest improvements on unseen environments, where long-horizon cues are critical for handling novel layouts.
These results demonstrate that the two objectives are complementary: SP enhances local spatial understanding, while TR strengthens temporal reasoning. Their combination yields a more reliable navigation policy and supports the design of our multi-task framework, reinforcing our goal of unifying spatial, temporal, and embodied reasoning within a single model.

\subsubsection{Effect of History Representation Strategies} 
\begin{table}[t]
\centering
\renewcommand{\arraystretch}{1.3}
\caption{Comparison of different history representation strategies on AerialVLN-S Val-Unseen split}
\label{tab:ablation_history}
\small
\begin{tabular*}{\linewidth}{@{\extracolsep{\fill}} l c c c}
\hline
History Strategy &  SR$\uparrow$ & SDTW$\uparrow$ \\
\hline
Current-Only $(2B)$  & 4.1 & 1.1 \\
Short-Term Memory Bank $(2B, k=8)$   & 4.3 & 1.2  \\
Long-Horizon Uniform Sampling $(2B, k=8)$  & 5.5  & 1.6 \\
Long-Horizon Uniform Sampling $(8B, k=8)$  & \textbf{5.9}  & \textbf{1.6} \\
\hline
\end{tabular*}
\end{table}

\begin{figure}[t!]
    \centering
    \color{black}
    \includegraphics[width= 0.95 \linewidth]{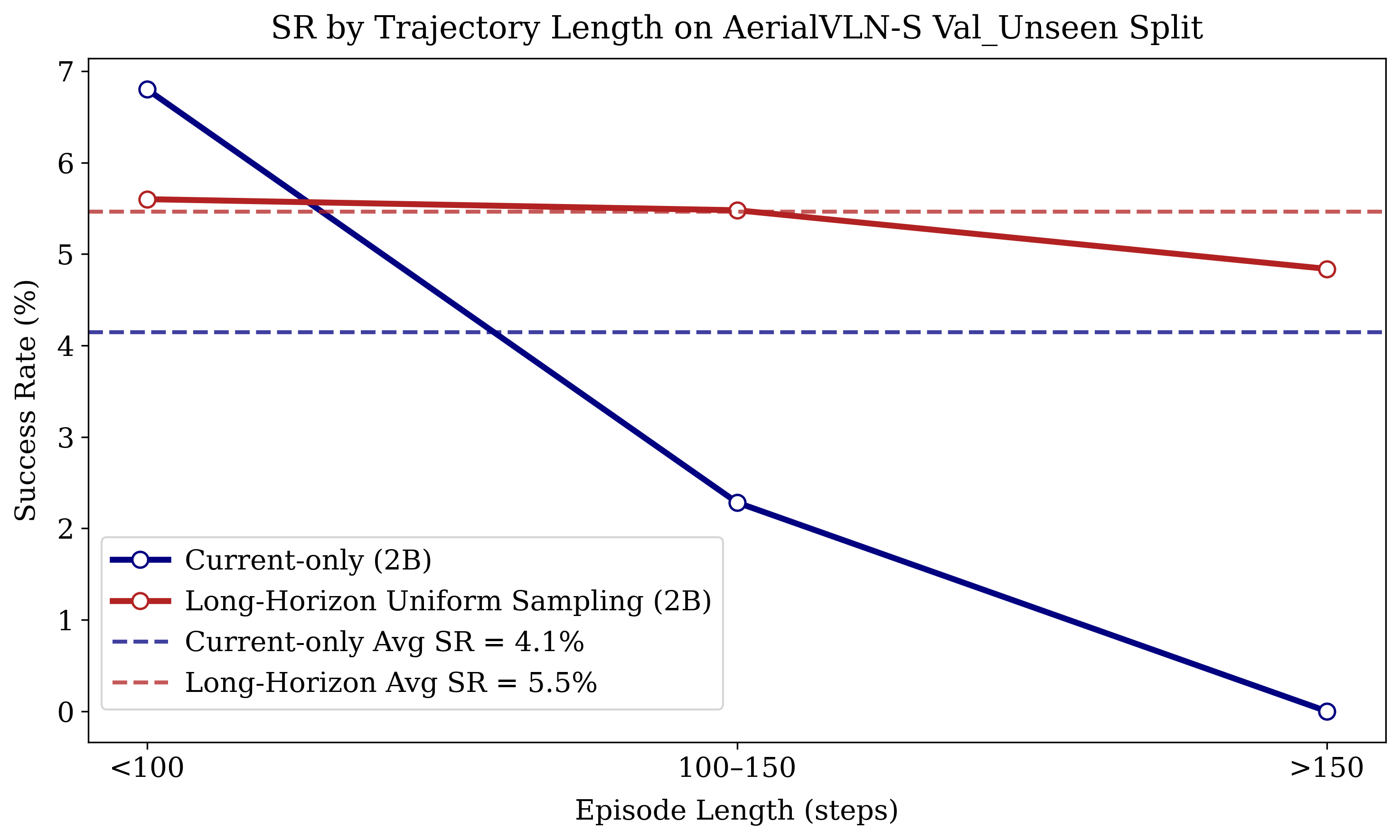}
    \caption{Success rate across different trajectory lengths on the AerialVLN-S Val-Unseen split. Long-Horizon Uniform Sampling maintains stable performance across episodes of varying lengths.
    }
    \label{fig:subfig1}
\end{figure}

We compare three history representation strategies.
(1) \textit{Current-Only}, where the model relies solely on the latest observation, often leading to unstable behaviors such as repeated fine-grained heading corrections.
(2) \textit{Short-Term Memory Bank}, which maintains a fixed-size memory bank following a first-in-first-out (FIFO) policy, keeping only the most recent observations and discarding earlier ones \cite{gao2025openfly}. This provides short-range temporal cues while discarding earlier information essential for global navigation.
(3) \textit{Long-Horizon Uniform Sampling}, which uniformly samples a small set of frames from the entire past trajectory to provide long-range visual context. Table~\ref{tab:ablation_history}  compares these history representation strategies. Among them, Long-Horizon Uniform Sampling ($k=8$) achieves the best overall performance. By drawing observations uniformly from the entire past trajectory, this strategy preserves both short-term motion cues and long-range scene information. As a result, it provides a more reliable temporal context than relying solely on the current frame or restricting history to a short FIFO queue.  We compare these strategies using the NVILA-2B model for computational efficiency. The larger 8B variant follows the same trend and achieves consistently higher accuracy. To further analyze how different history strategies behave under varying temporal horizons, we group the Val-Unseen episodes into three categories based on their trajectory length and report the per-step success rate for each method in Fig. \ref{fig:subfig1}. The Current-Only baseline performs reasonably well on short trajectories because the goal is often visible early, but its accuracy declines sharply as the trajectory becomes longer, indicating limited robustness without historical context.
The Long-Horizon Uniform Sampling strategy maintains higher performance across all trajectory lengths, with clear advantages on medium and long episodes where global scene information becomes crucial. These results highlight the importance of accessing long-range temporal cues for stable navigation in large, visually diverse environments.

\begin{table}[t]
\centering
\renewcommand{\arraystretch}{1.3}
\caption{Ablation study on the effect of action merging (AM), keyframe selection (KS) and label reweighting (LR) on AerialVLN-S Val-Seen split. 
All of them consistently improve performance across all metrics}
\label{tab:ablation_action_keyframe}
\small
\begin{tabular*}{0.95 \linewidth}{@{\extracolsep{\fill}} c c c | c c c c}
\hline
AM & KS & LR & NE$\downarrow$ & SR$\uparrow$ & OSR$\uparrow$ & SDTW$\uparrow$ \\
\hline
 &      &  & 90.2 & 1.8 & 20.5 & 0.7 \\
\checkmark &      & & 91.5 & 3.0 & 26.8 & 1.4 \\
\checkmark &  \checkmark    &  & 87.1 & 5.7 & 28.9 & 1.9 \\
\checkmark &   \checkmark    & \checkmark & \textbf{82.1} & \textbf{8.1} & \textbf{32.2} & \textbf{3.8} \\
\hline
\end{tabular*}

\end{table}

\subsubsection{Effect of Action Merging, Keyframe Selection, and Label Reweighting} Table~\ref{tab:ablation_action_keyframe} reports the effect of action merging (AM), keyframe selection (KS), and label reweighting (LR).
For computational efficiency, this ablation uses the NVILA-2B model without auxiliary tasks, and all variants adopt a long-horizon uniform sampling of 8 history frames. 
Enabling action merging leads to immediate gains, as consolidating micro-actions into coherent motion segments produces cleaner supervisory signals compared to raw steps.
Adding keyframe selection further improves navigation accuracy. Selecting the boundaries of merged actions as keyframes provides a more structured trajectory history in which visual transitions align with meaningful motion changes. This reduces temporal redundancy while preserving observations most relevant to spatial landmarks.
Finally, incorporating label reweighting achieves the best overall performance. By balancing the contribution of long merged segments and short corrective actions, our design prevents the model from overfitting to frequent yet trivial motions, thereby strengthening action-level supervision.  

We further ablate the maximum number of consecutive actions merged (i.e., the merge step) and report the results in Table~\ref {tab:ablation_merge_step}. As shown, a merge step of 3 achieves the best performance, while a merge step of 1 (i.e., no merging) retains the original fragmented action sequence and leads to lower navigation performance. Excessively large merge steps (e.g., 6) slightly degrade performance, likely because they skip important visual cues and reduce the model’s ability to adapt to local changes. We therefore set the maximum merge length to 3. Keyframes are then selected at the boundaries of these refined segments, ensuring sufficient temporal coverage and preserving landmarks that may appear along extended straight paths.
Overall, the combined use of AM, KS, and LR yields substantial improvements in SR and SDTW, demonstrating that structuring the action sequence and rebalancing its supervision are both crucial for reliable aerial navigation.

\begin{table}[t]
\color{black}
\centering
\renewcommand{\arraystretch}{1.3}
\caption{Ablation on the maximum number of consecutive actions merged evaluated on AerialVLN‑S val‑seen split. The best performance is achieved with a merge step of 3}
\label{tab:ablation_merge_step}
\small
\begin{tabular*}{0.95\linewidth}{@{\extracolsep{\fill}} c  c c c c}
\hline
Merge Step & NE$\downarrow$ & SR$\uparrow$ & OSR$\uparrow$ & SDTW$\uparrow$ \\
\hline
1  & 90.2 & 1.8 & 20.5 & 0.7 \\
\textbf{3} & \textbf{87.1} & \textbf{5.7} & \textbf{28.9} & \textbf{1.9} \\
6 & 89.5 & 5.1 & 25.3 & 1.4 \\
\hline
\end{tabular*}
\end{table}
\begin{table}[t]

\color{black}
\centering
\renewcommand{\arraystretch}{1.3}
\caption{Quantitative failure case analysis on the AerialVLN-S val-unseen split  
Note that $\dagger$ indicates our method without auxiliary tasks.
$\Delta$ denotes the difference in episode counts between the proposed model and the variant trained without auxiliary tasks.}
\label{tab:failure_analysis}
\small
\begin{tabular*}{0.95\linewidth}{@{\extracolsep{\fill}} l c c c}
\hline
Failure Type & Ours & Ours$^\dagger$ & $\Delta$ \\
\hline 
Perception-related & 194 & 204 & $-10$ \\
Long-horizon drift & 124 & 135 & $-11$ \\
Collision & 142 & 150 & $-8$ \\
Stop failure & 107 & 90 & $+17$ \\
\hline
\end{tabular*}
\end{table}

\subsection{Failure Case Analysis}
Although our method achieves strong performance and 
outperforms competitive baselines, it is still susceptible to several typical 
failure modes. To better understand the remaining challenges and guide future 
improvements, we further analyze the most common failure cases below. A common failure mode arises from ambiguous instructions that provide only a sequence of low-level actions without any landmark references 
(e.g., “rotate, go straight, turn left, turn right, hold on”).  The absence of such grounding references makes it difficult for the agent to determine where in the environment each action should be executed, leading to misalignment between the instruction and the visual context. We manually identify 21 severely underspecified instructions whose action sequences lack explicit landmark references. On this subset, both the proposed model and the variant without auxiliary tasks fail. This observation is consistent with previous work\cite{zhang2025citynavagent} as the absence of explicit landmark references prevents the agent from grounding the action sequence to specific locations in the environment, leading to inherently unreliable decision-making. Excluding these cases, we analyze the remaining episodes using trajectory-level metrics and conduct a quantitative analysis on the AerialVLN-S val-unseen split. Failure cases are categorized into stop failures (OSR=1 with incorrect termination), collision failures (OSR=0 with obstacle contact), long-horizon drift (OSR=0 with low trajectory similarity measured by nDTW), and perception-related failures (remaining cases exhibiting partial trajectory alignment but incorrect intermediate grounding). As summarized in Table~\ref{tab:failure_analysis}, the auxiliary tasks reduce both long-horizon drift and perception-related failures compared to the model trained without auxiliary supervision. This suggests improved global alignment over extended trajectories and more stable visual grounding under monocular observation. Collision failures are slightly reduced. In contrast, stop failures increase, indicating that the auxiliary tasks enable the agent to reach the goal region more frequently while termination decisions remain imperfect. Overall, the results show that the auxiliary tasks mitigate specific failure modes, particularly long-horizon drift and perception-related errors, by strengthening alignment between visual perception, textual language, and embodied action.

\section{Conclusion and Discussion}
In this paper, we present a unified framework for aerial vision-and-language navigation that operates solely on monocular egocentric RGB inputs and natural language instructions, eliminating the need for panoramic observations, depth sensing, or external localization modules. To address core challenges posed by the complexity of large-scale outdoor scenes and the long-horizon nature of aerial navigation,  we reformulate aerial VLN as a next-token prediction problem and integrate spatial perception, trajectory reasoning, and action generation through task-specific prompts. Experiments on the AerialVLN and OpenFly benchmarks demonstrate that our method achieves new state-of-the-art performance among RGB-only approaches and significantly narrows the gap with panoramic RGB-D counterparts. This work highlights the potential of scalable, prompt-driven multimodal learning as an efficient solution for deploying embodied aerial agents in real-world scenarios.

Despite the encouraging results, several limitations remain. First, the current framework relies on passive monocular observations, which may lead to incomplete environmental coverage under restricted field-of-view settings. Second, performance can degrade under ambiguous instructions that lack explicit landmark references. Third, long-horizon trajectory alignment remains challenging when small prediction errors accumulate over time. Future work will explore active viewpoint control to expand perceptual coverage, ambiguity-aware instruction clarification to improve grounding under underspecified commands, and hierarchical or staged planning strategies to enhance long-range consistency. Moreover, the aerial VLN method could benefit from lightweight obstacle-aware planning modules that operate alongside the navigation policy. Further improvements in inference efficiency may be achieved through more advanced acceleration techniques and hardware-aware optimizations toward deployment on resource-constrained UAV platforms.


{\appendices

\section{Additional Implementation Details}
\subsection{Action Parsing Procedure}
\label{app:action_parser}
For reproducibility, we provide additional details of the action parsing procedure, which maps generated textual actions to executable action sequences. The parser first matches the generated text to one predefined action type using regular expressions. It then extracts the numeric argument, when present, and decomposes the command into repeated primitive actions according to the predefined unit of the matched action type. If no numeric argument is present, a single primitive action of the matched type is returned. If no valid action type can be identified, a conservative fallback action is adopted.

\subsection{Trajectory-Reasoning Annotation Protocol}
\label{app:tr_annotation}

Additional details of the trajectory-reasoning annotation process are provided below. The annotation process followed four explicit rules:
\begin{itemize}
    \item \textit{Matching Principle}: each sub-trajectory had to match a semantically coherent sub-instruction.
    \item \textit{Clear End Point Principle}: each sub-instruction had to contain a clearly described endpoint.
    \item \textit{Length Constraint}: each sub-trajectory was constrained to span approximately 4--16 keyframes after action merging to maintain temporal coherence without introducing excessive granularity.
    \item \textit{Instruction Source Constraint}: the sub-instructions were directly extracted from the original instruction without paraphrasing or linguistic rewriting.
\end{itemize}

The trajectory-reasoning supervision was constructed exclusively from the training split and further expanded through programmatic composition. Severely ambiguous instructions, mainly those without a clear landmark reference or identifiable endpoint, were excluded during annotation. Additional training instances were then constructed programmatically by concatenating 2--3 consecutive annotated sub-trajectories and their corresponding sub-instructions, without requiring further human labeling. An illustrative annotation example is provided in Table~\ref{tab:tr_annotation_example}.

\begin{table*}[t]
\centering
\caption{Illustrative example of manual trajectory segmentation used for trajectory-reasoning annotation}
\label{tab:tr_annotation_example}
\small
\setlength{\tabcolsep}{4pt}
\renewcommand{\arraystretch}{1.3}
\begin{tabular}{p{2.6cm}|p{2.8cm}|p{11.0cm}}
\hline
\textbf{Trajectory unit} & \textbf{Field} & \textbf{Content} \\
\hline
Trajectory & Original instruction & Take off to first floor height and turn right by $180^\circ$ then proceed. Stop at bridge. Turn right by $90^\circ$ then ascend to roof height. Turn right by $180^\circ$ and proceed towards yellow building. Proceed towards satellites on roof. \\
\hline
Sub-trajectory \#1 & Sub-instruction & Take off to first floor height and turn right by $180^\circ$ then proceed. Stop at bridge. \\
\cline{2-3}
 & Keyframe timesteps & 0, 2, 5, 8, 11, 13, 16, 18, 19, 22, 24, 25, 27 \\
\hline
Sub-trajectory \#2 & Sub-instruction & Turn right by $90^\circ$ then ascend to roof height. Turn right by $180^\circ$ and proceed towards yellow building. Proceed towards satellites on roof. \\
\cline{2-3}
 & Keyframe timesteps & 27, 30, 33, 34, 37, 40, 43, 44, 47, 50, 53, 54, 55, 58, 59, 60 \\
\hline
\end{tabular}
\end{table*}

\begin{table}[t]
\centering
\renewcommand{\arraystretch}{1.3}
\caption{Per-scene trajectory statistics of the OpenFly-S benchmark}
\label{tab:openflys_scene_stats}
\begin{tabular*}{0.95\linewidth}{@{\extracolsep{\fill}} l c c}
\hline
Scene & Train Split & Test-Seen Split \\
\hline
env\_airsim\_16 & 9,556 & 202 \\
env\_airsim\_18 & 9,731 & 203 \\
env\_airsim\_23 & 3,429 & 200 \\
env\_airsim\_26 & 9,690 & 202 \\
env\_airsim\_sh & 18,842 & 202 \\
env\_airsim\_gz & 9,756 & 201 \\
\hline
Total & 61,004 & 1,210 \\
\hline
\end{tabular*}
\end{table}

\subsection{OpenFly-S Benchmark Construction}
\label{app:openflys}
The OpenFly-S benchmark is constructed by applying a deterministic scene-level filtering rule to the original OpenFly benchmark. Specifically, we retain only the trajectories belonging to six UE4 scenes. No additional trajectory-level manual selection, heuristic resampling, or performance-based filtering is applied beyond this scene-level restriction. The split protocol of OpenFly-S is inherited directly from the original OpenFly split definition and then filtered by scene identity. Concretely, the training split of OpenFly-S contains all training trajectories from the six selected scenes, while the test-seen split contains all seen-test trajectories from the same scenes. Therefore, once the scene list is fixed, the OpenFly-S subset can be reconstructed exactly from the original OpenFly data. The resulting per-scene trajectory statistics are reported in Table~\ref{tab:openflys_scene_stats}. In total, OpenFly-S contains 61,004 trajectories in the training split and 1,210 trajectories in the test-seen split.

\section{Additional Experimental Results}
\subsection{Multi-Seed Results}
\label{app:multiseed}
We further report multi-seed results to examine the stability of the proposed method across different training seeds. Training was repeated with three random seeds, and we report the mean and standard deviation. The evaluation follows the same setting as in the main paper and is conducted on the AerialVLN-S benchmark \cite{liu2023aerialvln} under the validation seen split. Results are provided for both the full model and the variant trained without auxiliary tasks. For reference, we also include the originally reported single-run results of representative RGB-only VLM-based baselines, namely NaVid \cite{zhang2024navid} and OpenFly \cite{gao2025openfly}. Since multi-seed statistics are not available for these baselines, they are retained in their original reporting format. The corresponding results are summarized in Table~\ref{tab:multiseed_results}. These results indicate that the proposed method maintains stable improvements over the variant without auxiliary tasks across different training seeds, while remaining competitive with the representative baselines.

\begin{table}[t]
\renewcommand{\arraystretch}{1.3}
\centering
\caption{Multi-seed results over three training random seeds, reported as mean $\pm$ standard deviation. Note that $\dagger$ indicates the variant trained without auxiliary tasks}
\begin{tabular*}{0.95\linewidth}{@{\extracolsep{\fill}}lcccc}
\hline
 Method & NE$\downarrow$ & SR$\uparrow$ & OSR$\uparrow$  & SDTW$\uparrow$  \\
\hline
NaVid \cite{zhang2024navid} & 105.1  & 6.8 & 15.5 & 1.1  \\
OpenFly \cite{gao2025openfly} & 127.2 & 8.1 & 21.8 & 1.6  \\
Ours$^\dagger$ & \underline{84.6} {\color{gray}{\scriptsize($\pm$1.7)}} & \underline{9.9} {\color{gray}{\scriptsize($\pm$1.1)}} & \underline{34.6} {\color{gray}{\scriptsize($\pm$5.2)}} & \underline{5.1} {\color{gray}{\scriptsize($\pm$0.7)}} \\
Ours & \textbf{76.8} {\color{gray}{\scriptsize($\pm$2.5)}} & \textbf{12.3} {\color{gray}{\scriptsize($\pm$1.1)}} & \textbf{38.9} {\color{gray}{\scriptsize($\pm$1.0)}} & \textbf{6.5} {\color{gray}{\scriptsize($\pm$1.0)}} \\

\hline
\end{tabular*}
\label{tab:multiseed_results}
\end{table}

}

\bibliographystyle{IEEEtran}
\bibliography{ref}

@inproceedings{krantz2020cma,
  title={Beyond the nav-graph: Vision-and-language navigation in continuous environments},
  author={Krantz, Jacob and Wijmans, Erik and Majumdar, Arjun and Batra, Dhruv and Lee, Stefan},
  booktitle={European Conference on Computer Vision},
  pages={104--120},
  year={2020},
  organization={Springer}
}

@inproceedings{liu2023aerialvln,
  title={Aerialvln: Vision-and-language navigation for uavs},
  author={Liu, Shubo and Zhang, Hongsheng and Qi, Yuankai and Wang, Peng and Zhang, Yanning and Wu, Qi},
  booktitle={Proceedings of the IEEE/CVF International Conference on Computer Vision},
  pages={15384--15394},
  year={2023}
}

@article{gao2024aerialstmr,
  title={Aerial vision-and-language navigation via semantic-topo-metric representation guided LLM reasoning},
  author={Gao, Yunpeng and Wang, Zhigang and Jing, Linglin and Wang, Dong and Li, Xuelong and Zhao, Bin},
  journal={arXiv preprint arXiv:2410.08500},
  year={2024}
}

@inproceedings{zhang2025citynavagent,
    title = "{C}ity{N}av{A}gent: Aerial Vision-and-Language Navigation with Hierarchical Semantic Planning and Global Memory",
    author = "Zhang, Weichen  and
      Gao, Chen  and
      Yu, Shiquan  and
      Peng, Ruiying  and
      Zhao, Baining  and
      Zhang, Qian  and
      Cui, Jinqiang  and
      Chen, Xinlei  and
      Li, Yong",
    booktitle = "Proceedings of the 63rd Annual Meeting of the Association for Computational Linguistics (Volume 1: Long Papers)",
    year = "2025",
    publisher = "Association for Computational Linguistics",
    doi = "10.18653/v1/2025.acl-long.1511",
    pages = "31292--31309",
    ISBN = "979-8-89176-251-0",
    
}

@article{gao2025openfly,
  title={OpenFly: A versatile toolchain and large-scale benchmark for aerial vision-language navigation},
  author={Gao, Yunpeng and Li, Chenhui and You, Zhongrui and Liu, Junli and Li, Zhen and Chen, Pengan and Chen, Qizhi and Tang, Zhonghan and Wang, Liansheng and Yang, Penghui and others},
  journal={arXiv e-prints},
  pages={arXiv--2502},
  year={2025}
}

@article{zhang2024navid,
        title={NaVid: Video-based VLM Plans the Next Step for Vision-and-Language Navigation},
        author={Zhang, Jiazhao and Wang, Kunyu and Xu, Rongtao and Zhou, Gengze and Hong, Yicong and Fang, Xiaomeng and Wu, Qi and Zhang, Zhizheng and Wang, He},
        journal={Robotics: Science and Systems},
        year={2024}
      }

@article{zhao2025aerialviewselection,
  title={Aerial Vision-and-Language Navigation with Grid-based View Selection and Map Construction},
  author={Zhao, Ganlong and Li, Guanbin and Pan, Jia and Yu, Yizhou},
  journal={arXiv preprint arXiv:2503.11091},
  year={2025}
}

@inproceedings{lee2025citynavdataset,
  title={CityNav: A Large-Scale Dataset for Real-World Aerial Navigation},
  author={Lee, Jungdae and Miyanishi, Taiki and Kurita, Shuhei and Sakamoto, Koya and Azuma, Daichi and Matsuo, Yutaka and Inoue, Nakamasa},
  booktitle={Proceedings of the IEEE/CVF International Conference on Computer Vision},
  pages={5912--5922},
  year={2025}
}

@article{zhang2025open3dvqa,
  title={Open3dvqa: A benchmark for comprehensive spatial reasoning with multimodal large language model in open space},
  author={Zhang, Weichen and Zhou, Zile and Zheng, Zhiheng and Gao, Chen and Cui, Jinqiang and Li, Yong and Chen, Xinlei and Zhang, Xiao-Ping},
  journal={arXiv preprint arXiv:2503.11094},
  year={2025}
}

@inproceedings{hudson2019gqa,
  title={Gqa: A new dataset for real-world visual reasoning and compositional question answering},
  author={Hudson, Drew A and Manning, Christopher D},
  booktitle={Proceedings of the IEEE/CVF conference on computer vision and pattern recognition},
  pages={6700--6709},
  year={2019}
}

@inproceedings{anderson2018r2r,
  title={Vision-and-language navigation: Interpreting visually-grounded navigation instructions in real environments},
  author={Anderson, Peter and Wu, Qi and Teney, Damien and Bruce, Jake and Johnson, Mark and S{\"u}nderhauf, Niko and Reid, Ian and Gould, Stephen and Van Den Hengel, Anton},
  booktitle={Proceedings of the IEEE conference on computer vision and pattern recognition},
  pages={3674--3683},
  year={2018}
}

@inproceedings{ku2020rxr,
  title={Room-Across-Room: Multilingual Vision-and-Language Navigation with Dense Spatiotemporal Grounding},
  author={Ku, Alexander and Anderson, Peter and Patel, Roma and Ie, Eugene and Baldridge, Jason},
  booktitle={Proceedings of the 2020 Conference on Empirical Methods in Natural Language Processing (EMNLP)},
  pages={4392--4412},
  year={2020}
}

@inproceedings{hong2021vln_bert,
  title={Vln bert: A recurrent vision-and-language bert for navigation},
  author={Hong, Yicong and Wu, Qi and Qi, Yuankai and Rodriguez-Opazo, Cristian and Gould, Stephen},
  booktitle={Proceedings of the IEEE/CVF conference on Computer Vision and Pattern Recognition},
  pages={1643--1653},
  year={2021}
}

@article{wang2025monodream,
  title={Monodream: Monocular vision-language navigation with panoramic dreaming},
  author={Wang, Shuo and Wang, Yongcai and Li, Wanting and Wang, Yucheng and Chen, Maiyue and Wang, Kaihui and Su, Zhizhong and Cai, Xudong and Jin, Yeying and Li, Deying and others},
  journal={arXiv preprint arXiv:2508.02549},
  year={2025}
}

@article{zhang2024uninavid,
  title={Uni-navid: A video-based vision-language-action model for unifying embodied navigation tasks},
  author={Zhang, Jiazhao and Wang, Kunyu and Wang, Shaoan and Li, Minghan and Liu, Haoran and Wei, Songlin and Wang, Zhongyuan and Zhang, Zhizheng and Wang, He},
  journal={arXiv preprint arXiv:2412.06224},
  year={2024}
}

@article{wu2024vln_survey,
  title={Vision-language navigation: a survey and taxonomy},
  author={Wu, Wansen and Chang, Tao and Li, Xinmeng and Yin, Quanjun and Hu, Yue},
  journal={Neural Computing and Applications},
  volume={36},
  number={7},
  pages={3291--3316},
  year={2024},
  publisher={Springer}
}

@inproceedings{wang2023dreamwalker,
  title={Dreamwalker: Mental planning for continuous vision-language navigation},
  author={Wang, Hanqing and Liang, Wei and Van Gool, Luc and Wang, Wenguan},
  booktitle={Proceedings of the IEEE/CVF international conference on computer vision},
  pages={10873--10883},
  year={2023}
}

@article{an2024etpnav,
  title={Etpnav: Evolving topological planning for vision-language navigation in continuous environments},
  author={An, Dong and Wang, Hanqing and Wang, Wenguan and Wang, Zun and Huang, Yan and He, Keji and Wang, Liang},
  journal={IEEE Transactions on Pattern Analysis and Machine Intelligence},
  year={2024},
  publisher={IEEE}
}

@inproceedings{wang2023gridmm,
  title={Gridmm: Grid memory map for vision-and-language navigation},
  author={Wang, Zihan and Li, Xiangyang and Yang, Jiahao and Liu, Yeqi and Jiang, Shuqiang},
  booktitle={Proceedings of the IEEE/CVF International conference on computer vision},
  pages={15625--15636},
  year={2023}
}

@article{qi2025vln,
  title={VLN-R1: Vision-Language Navigation via Reinforcement Fine-Tuning},
  author={Qi, Zhangyang and Zhang, Zhixiong and Yu, Yizhou and Wang, Jiaqi and Zhao, Hengshuang},
  journal={arXiv preprint arXiv:2506.17221},
  year={2025}
}

@inproceedings{cheng2025navila,
        title={Navila: Legged robot vision-language-action model for navigation},
        author={Cheng, An-Chieh and Ji, Yandong and Yang, Zhaojing and Gongye, Zaitian and Zou, Xueyan and Kautz, Jan and B{\i}y{\i}k, Erdem and Yin, Hongxu and Liu, Sifei and Wang, Xiaolong},
        booktitle={RSS},
        year={2025}
}

@inproceedings{wangopenuav,
  title={Towards Realistic UAV Vision-Language Navigation: Platform, Benchmark, and Methodology},
  author={Wang, Xiangyu and Yang, Donglin and Wang, Ziqin and Kwan, Hohin and Chen, Jinyu and Wu, Wenjun and Li, Hongsheng and Liao, Yue and Liu, Si},
  booktitle={The Thirteenth International Conference on Learning Representations}
}

@inproceedings{fan2023AVDN,
  title={Aerial vision-and-dialog navigation},
  author={Fan, Yue and Chen, Winson and Jiang, Tongzhou and Zhou, Chun and Zhang, Yi and Wang, Xin},
  booktitle={Findings of the Association for Computational Linguistics: ACL 2023},
  pages={3043--3061},
  year={2023}
}

@article{lindqvist2021obstacleavoidance,
  title={Reactive navigation of an unmanned aerial vehicle with perception-based obstacle avoidance constraints},
  author={Lindqvist, Bj{\"o}rn and Mansouri, Sina Sharif and Halu{\v{s}}ka, Jakub and Nikolakopoulos, George},
  journal={IEEE Transactions on Control Systems Technology},
  volume={30},
  number={5},
  pages={1847--1862},
  year={2021},
  publisher={IEEE}
}

@article{zhang2025obstacleavoidancenature,
  title={Learning vision-based agile flight via differentiable physics},
  author={Zhang, Yuang and Hu, Yu and Song, Yunlong and Zou, Danping and Lin, Weiyao},
  journal={Nature Machine Intelligence},
  pages={1--13},
  year={2025},
  publisher={Nature Publishing Group UK London}
}

@article{wang2019targettracking,
  title={Development of UAV-based target tracking and recognition systems},
  author={Wang, Shuaijun and Jiang, Fan and Zhang, Bin and Ma, Rui and Hao, Qi},
  journal={IEEE Transactions on Intelligent Transportation Systems},
  volume={21},
  number={8},
  pages={3409--3422},
  year={2019},
  publisher={IEEE}
}

@article{savkin2023effective,
  title={Effective UAV navigation for cellular-assisted radio sensing, imaging, and tracking},
  author={Savkin, Andrey V and Ni, Wei and Eskandari, Mohsen},
  journal={IEEE Transactions on Vehicular Technology},
  volume={72},
  number={10},
  pages={13729--13733},
  year={2023},
  publisher={IEEE}
}

@article{khan2022uavsurvey,
  title={Emerging UAV technology for disaster detection, mitigation, response, and preparedness},
  author={Khan, Amina and Gupta, Sumeet and Gupta, Sachin Kumar},
  journal={Journal of Field Robotics},
  volume={39},
  number={6},
  pages={905--955},
  year={2022},
  publisher={Wiley Online Library}
}

@article{zhao2025cityeqa,
  title={Cityeqa: A hierarchical llm agent on embodied question answering benchmark in city space},
  author={Zhao, Yong and Xu, Kai and Zhu, Zhengqiu and Hu, Yue and Zheng, Zhiheng and Chen, Yingfeng and Ji, Yatai and Gao, Chen and Li, Yong and Huang, Jincai},
  journal={arXiv preprint arXiv:2502.12532},
  year={2025}
}

@inproceedings{wu2025aeroduo,
  title={AeroDuo: Aerial Duo for UAV-based Vision and Language Navigation},
  author={Wu, Ruipu and Zhang, Yige and Chen, Jinyu and Huang, Linjiang and Zhang, Shifeng and Zhou, Xu and Wang, Liang and Liu, Si},
  booktitle={Proceedings of the 33rd ACM International Conference on Multimedia},
  pages={2576--2585},
  year={2025}
}

@inproceedings{chen2024mapgpt,
  title={MapGPT: Map-Guided Prompting with Adaptive Path Planning for Vision-and-Language Navigation},
  author={Chen, Jiaqi and Lin, Bingqian and Xu, Ran and Chai, Zhenhua and Liang, Xiaodan and Wong, Kwan-Yee~K.},
  booktitle = "Proceedings of the 62nd Annual Meeting of the Association for Computational Linguistics",
  year={2024}
}

@article{achiam2023gpt,
  title={Gpt-4 technical report},
  author={Achiam, Josh and Adler, Steven and Agarwal, Sandhini and Ahmad, Lama and Akkaya, Ilge and Aleman, Florencia Leoni and Almeida, Diogo and Altenschmidt, Janko and Altman, Sam and Anadkat, Shyamal and others},
  journal={arXiv preprint arXiv:2303.08774},
  year={2023}
}

@article{leng2022pareto,
  title={Pareto refocusing for drone-view object detection},
  author={Leng, Jiaxu and Mo, Mengjingcheng and Zhou, Yinghua and Gao, Chenqiang and Li, Weisheng and Gao, Xinbo},
  journal={IEEE Transactions on Circuits and Systems for Video Technology},
  volume={33},
  number={3},
  pages={1320--1334},
  year={2022},
  publisher={IEEE}
}

@article{chen2023high,
  title={High-resolution feature pyramid network for small object detection on drone view},
  author={Chen, Zhaodong and Ji, Hongbing and Zhang, Yongquan and Zhu, Zhigang and Li, Yifan},
  journal={IEEE Transactions on Circuits and Systems for Video Technology},
  volume={34},
  number={1},
  pages={475--489},
  year={2023},
  publisher={IEEE}
}

@article{chen2023cross,
  title={Cross-drone transformer network for robust single object tracking},
  author={Chen, Guanlin and Zhu, Pengfei and Cao, Bing and Wang, Xing and Hu, Qinghua},
  journal={IEEE Transactions on Circuits and Systems for Video Technology},
  volume={33},
  number={9},
  pages={4552--4563},
  year={2023},
  publisher={IEEE}
}

@article{wu2024temporal,
  title={Temporal-spatial feature interaction network for multi-drone multi-object tracking},
  author={Wu, Han and Sun, Hao and Ji, Kefeng and Kuang, Gangyao},
  journal={IEEE Transactions on Circuits and Systems for Video Technology},
  year={2024},
  publisher={IEEE}
}

@ARTICLE{2024vln,
  author={Zhan, Zhaohuan and Qin, Jinghui and Zhuo, Wei and Tan, Guang},
  journal={IEEE Transactions on Circuits and Systems for Video Technology}, 
  title={Enhancing Vision and Language Navigation With Prompt-Based Scene Knowledge}, 
  year={2024},
  volume={34},
  number={10},
  pages={9745-9756},
  doi={10.1109/TCSVT.2024.3401451}}

@ARTICLE{2024planning,
  author={Chen, Bolei and Kang, Jiaxu and Zhong, Ping and Cui, Yongzheng and Lu, Siyi and Liang, Yixiong and Wang, Jianxin},
  journal={IEEE Transactions on Circuits and Systems for Video Technology}, 
  title={Think Holistically, Act Down-to-Earth: A Semantic Navigation Strategy With Continuous Environmental Representation and Multi-Step Forward Planning}, 
  year={2024},
  volume={34},
  number={5},
  pages={3860-3875},
  doi={10.1109/TCSVT.2023.3324380}}

@ARTICLE{NavComposer,
  author={He, Zongtao and Wang, Liuyi and Chen, Lu and Liu, Chengju and Chen, Qijun},
  journal={IEEE Transactions on Circuits and Systems for Video Technology}, 
  title={NavComposer: Composing Language Instructions for Navigation Trajectories through Action-Scene-Object Modularization}, 
  year={2025},
  volume={},
  number={},
  pages={1-1},
  doi={10.1109/TCSVT.2025.3596386}}

@inproceedings{misra2018mapping,
    title = "Mapping Instructions to Actions in 3{D} Environments with Visual Goal Prediction",
    author = "Misra, Dipendra  and
      Bennett, Andrew  and
      Blukis, Valts  and
      Niklasson, Eyvind  and
      Shatkhin, Max  and
      Artzi, Yoav",
    editor = "Riloff, Ellen  and
      Chiang, David  and
      Hockenmaier, Julia  and
      Tsujii, Jun{'}ichi",
    booktitle = "Proceedings of the 2018 Conference on Empirical Methods in Natural Language Processing",
    month = oct # "-" # nov,
    year = "2018",
    address = "Brussels, Belgium",
    publisher = "Association for Computational Linguistics",
    url = "https://aclanthology.org/D18-1287/",
    doi = "10.18653/v1/D18-1287",
    pages = "2667--2678"
}

@inproceedings{liu2025nvila,
  title={Nvila: Efficient frontier visual language models},
  author={Liu, Zhijian and Zhu, Ligeng and Shi, Baifeng and Zhang, Zhuoyang and Lou, Yuming and Yang, Shang and Xi, Haocheng and Cao, Shiyi and Gu, Yuxian and Li, Dacheng and others},
  booktitle={Proceedings of the Computer Vision and Pattern Recognition Conference},
  pages={4122--4134},
  year={2025}
}

@inproceedings{zhai2023sigclip,
  title={Sigmoid loss for language image pre-training},
  author={Zhai, Xiaohua and Mustafa, Basil and Kolesnikov, Alexander and Beyer, Lucas},
  booktitle={Proceedings of the IEEE/CVF international conference on computer vision},
  pages={11975--11986},
  year={2023}
}

@article{yang2024qwen2technicalreport,
    title   = {Qwen2 Technical Report}, 
    author  = {An Yang and Baosong Yang and Binyuan Hui and Bo Zheng and Bowen Yu and Chang Zhou and Chengpeng Li and Chengyuan Li and Dayiheng Liu and Fei Huang and Guanting Dong and Haoran Wei and Huan Lin and Jialong Tang and Jialin Wang and Jian Yang and Jianhong Tu and Jianwei Zhang and Jianxin Ma and Jin Xu and Jingren Zhou and Jinze Bai and Jinzheng He and Junyang Lin and Kai Dang and Keming Lu and Keqin Chen and Kexin Yang and Mei Li and Mingfeng Xue and Na Ni and Pei Zhang and Peng Wang and Ru Peng and Rui Men and Ruize Gao and Runji Lin and Shijie Wang and Shuai Bai and Sinan Tan and Tianhang Zhu and Tianhao Li and Tianyu Liu and Wenbin Ge and Xiaodong Deng and Xiaohuan Zhou and Xingzhang Ren and Xinyu Zhang and Xipin Wei and Xuancheng Ren and Yang Fan and Yang Yao and Yichang Zhang and Yu Wan and Yunfei Chu and Yuqiong Liu and Zeyu Cui and Zhenru Zhang and Zhihao Fan},
    journal = {arXiv preprint arXiv:2407.10671},
    year    = {2024}
}

@INPROCEEDINGS{Navid-4D,
  author={Liu, Haoran and Wan, Weikang and Yu, Xiqian and Li, Minghan and Zhang, Jiazhao and Zhao, Bo and Chen, Zhibo and Wang, Zhongyuan and Zhang, Zhizheng and Wang, He},
  booktitle={2025 IEEE International Conference on Robotics and Automation (ICRA)}, 
  title={Na Vid-4D: Unleashing Spatial Intelligence in Egocentric RGB-D Videos for Vision-and-Language Navigation}, 
  year={2025},
  volume={},
  number={},
  pages={10607-10615},
  doi={10.1109/ICRA55743.2025.11128467}}

@inproceedings{anderson2018seq2se,
  title={Vision-and-language navigation: Interpreting visually-grounded navigation instructions in real environments},
  author={Anderson, Peter and Wu, Qi and Teney, Damien and Bruce, Jake and Johnson, Mark and S{\"u}nderhauf, Niko and Reid, Ian and Gould, Stephen and Van Den Hengel, Anton},
  booktitle={Proceedings of the IEEE conference on computer vision and pattern recognition},
  pages={3674--3683},
  year={2018}
}

@article{lin2024awq,
  title={Awq: Activation-aware weight quantization for on-device llm compression and acceleration},
  author={Lin, Ji and Tang, Jiaming and Tang, Haotian and Yang, Shang and Chen, Wei-Ming and Wang, Wei-Chen and Xiao, Guangxuan and Dang, Xingyu and Gan, Chuang and Han, Song},
  journal={Proceedings of machine learning and systems},
  volume={6},
  pages={87--100},
  year={2024}
}

@inproceedings{uav-on,
author = {Xiao, Jianqiang and Sun, Yuexuan and Shao, Yixin and Gan, Boxi and Liu, Rongqiang and Wu, Yanjin and Guan, Weili and Deng, Xiang},
title = {UAV-ON: A Benchmark for Open-World Object Goal Navigation with Aerial Agents},
year = {2025},
isbn = {9798400720352},
doi = {10.1145/3746027.3758251},
booktitle = {Proceedings of the 33rd ACM International Conference on Multimedia},
pages = {13023–13029},
numpages = {7},

}

@article{wu2025vla-an,
  title={VLA-AN: An Efficient and Onboard Vision-Language-Action Framework for Aerial Navigation in Complex Environments},
  author={Wu, Yuze and Zhu, Mo and Li, Xingxing and Du, Yuheng and Fan, Yuxin and Li, Wenjun and Han, Zhichao and Zhou, Xin and Gao, Fei},
  journal={arXiv preprint arXiv:2512.15258},
  year={2025}
}

@inproceedings{cai2025flightgpt,
  title={Flightgpt: Towards generalizable and interpretable uav vision-and-language navigation with vision-language models},
  author={Cai, Hengxing and Dong, Jinhan and Tan, Jingjun and Deng, Jingcheng and Li, Sihang and Gao, Zhifeng and Wang, Haidong and Su, Zicheng and Sumalee, Agachai and Zhong, Renxin},
  booktitle={Proceedings of the 2025 Conference on Empirical Methods in Natural Language Processing},
  pages={6670--6687},
  year={2025}
}

\newpage
\begin{IEEEbiography}
[{\includegraphics[width=1in,height=1.25in,clip,keepaspectratio]{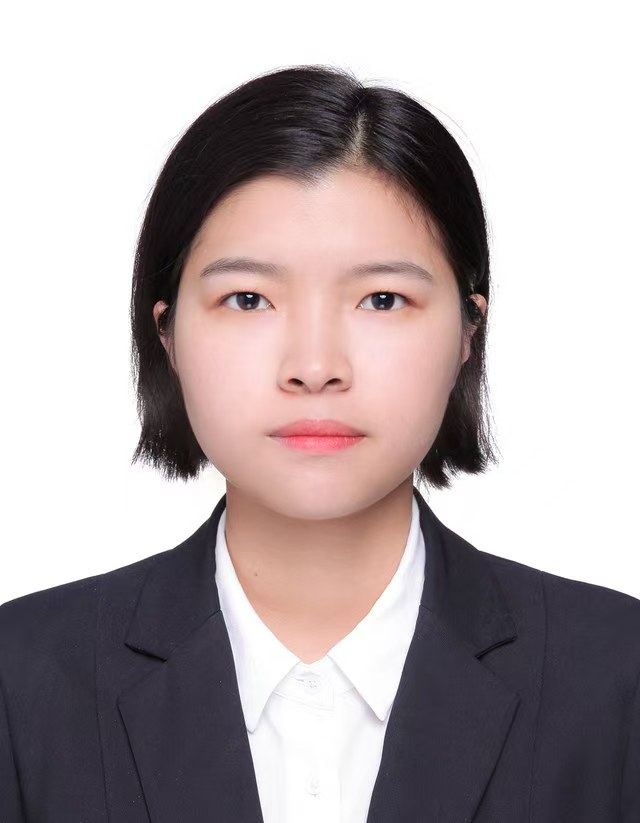}}]{Huilin Xu} (Student Member, IEEE) received the B.E. degree with honors from the School of Information Science and Engineering, Fudan University, Shanghai, China, in 2021. She is currently pursuing a Ph.D. degree at the Key Laboratory of Information Science of Electromagnetic Waves (MoE), Fudan University. Her research interests focus on visual reasoning, vision-based reinforcement learning, and robotic learning.
\end{IEEEbiography}

\begin{IEEEbiography}
[{\includegraphics[width=1in,height=1.25in,clip,keepaspectratio]{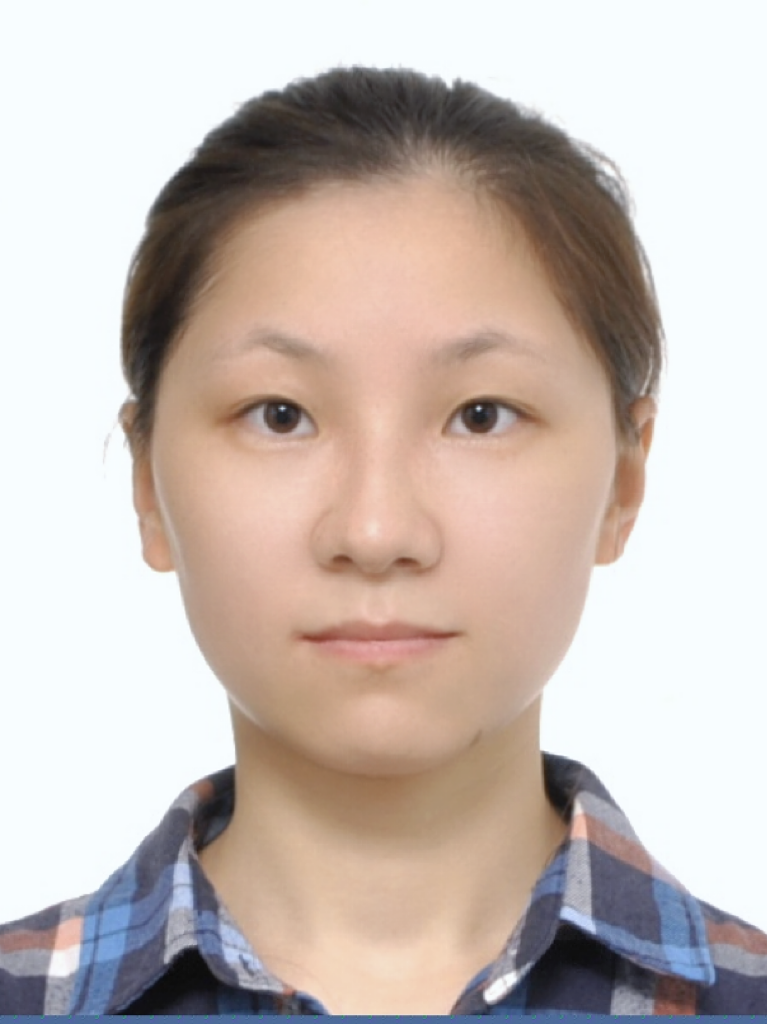}}]{Zhuoyang Liu}
    (Student Member, IEEE) received the B.E. degree from Wuhan University, Wuhan, Hubei, China, in 2020. She is currently pursuing the Ph.D. degree in electromagnetic science with the Key Laboratory of Information Science of Electromagnetic Waves, Fudan University, Shanghai, China. Her research interests include signal processing, Dual-Function Radar-Communication systems, reconfigurable intelligence surfaces, and the combination of artificial intelligence and electromagnetic waves.
\end{IEEEbiography}
\begin{IEEEbiography}
[{\includegraphics[width=1in,height=1.25in]{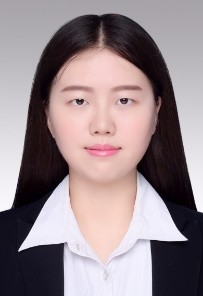}}]{Yixiang Luomei}
    (S’18-M’22) received the B.E. and M.E. in Electronic and Communication Engineering from Xidian University, Xi’an, China, and the Ph.D in Electronic Information Engineering from Fudan University, Shanghai, China, in 2014, 2018 and 2022, respectively. From 2022 to 2023, she was a Research Fellow with the National University of Singapore, Singapore. Since 2013, she has been a Research Assistant of the MoE Key Lab for Information Science of Electromagnetic Waves, School of Information Science and Technology, Fudan University, Shanghai, China. She has published more than 10 papers in peer-reviewed journals, among some conference papers and patents. She is currently working on intelligent processing of electromagnetic signals, UAV SAR imaging, and ISAC sensing.  
\end{IEEEbiography}

\begin{IEEEbiography}
[{\includegraphics[width=1in,height=1.25in,clip,keepaspectratio]{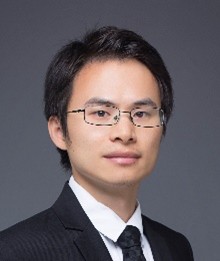}}]{Feng Xu}
    (Member, IEEE) received the B.E. degree (Hons.) in information engineering from Southeast University, Nanjing, China, in 2003, and the Ph.D. degree (Hons.) in electronic engineering from Fudan University, Shanghai, China, in 2008. From 2008 to 2010, he was a Post-Doctoral Fellow with the National Oceanic and Atmospheric Administration (NOAA) Center for Satellite Applications and Research, Camp Springs, MD, USA. From 2010 to 2013, he worked with Intelligent Automation Inc., Rockville, MD, USA, and NASA Goddard Space Flight Center, Greenbelt, MD, USA, as a Research Scientist. In 2012, he was selected for China’s Global Experts Recruitment Program and subsequently returned to Fudan University, in 2013, where he is currently a Professor and the Vice Dean of the School of Information Science and Technology and the Vice Director of the electromagnetic waves (MoE) Key Laboratory for Information Science of Electromagnetic Waves. 
His research interests include electromagnetic scattering modeling, SAR information retrieval, and radar system development. 

Dr. Xu is a Topic Associate Editor of the IEEE TRANSACTIONS OF GEOSCIENCE AND REMOTE SENSING and was an Associate Editor of the IEEE GEOSCIENCE AND REMOTE SENSING LETTERS (2014-2021). He is the Founding Chair of the IEEE GRSS Shanghai Chapter and an IEEE GRSS AdCom Member.

\end{IEEEbiography}

 




\vfill

\end{document}